\documentclass{clv3}

\usepackage{hyperref}
\usepackage{xcolor}
\definecolor{darkblue}{rgb}{0, 0, 0.5}
\hypersetup{colorlinks=true,citecolor=darkblue, linkcolor=darkblue, urlcolor=darkblue}
\usepackage{booktabs}
\usepackage{algorithm}
\usepackage{algorithmic}  
\usepackage{multirow}
\usepackage{bm}
\usepackage{amsmath}
\usepackage{arydshln}
\usepackage{multirow}
\usepackage{multicol}
\usepackage{arydshln}
\usepackage{graphicx}  

\issue{1}{1}{2020}

\runningtitle{Syntax Role for Neural Semantic Role Labeling}

\runningauthor{Li et al.}

\begin{document}

\title{Syntax Role for \\Neural Semantic Role Labeling}

\author{Zuchao Li}
\affil{Shanghai Jiao Tong University \\ Department of Computer Science and Engineering \\ \tt charlee@sjtu.edu.cn}

\author{Hai Zhao\thanks{Corresponding author.}}
\affil{Shanghai Jiao Tong University \\ Department of Computer Science and Engineering \\ \tt zhaohai@cs.sjtu.edu.cn}

\author{Shexia He}
\affil{Shanghai Jiao Tong University \\ Department of Computer Science and Engineering \\ \tt heshexia@sjtu.edu.cn}

\author{Jiaxun Cai}
\affil{Shanghai Jiao Tong University \\ Department of Computer Science and Engineering \\ \tt caijiaxun@sjtu.edu.cn}

\maketitle

\begin{abstract}
Semantic role labeling (SRL) is dedicated to recognizing the semantic predicate-argument structure of a sentence. 
Previous studies in terms of traditional models have shown syntactic information can make remarkable contributions to SRL performance; however, the necessity of syntactic information was challenged by a few recent neural SRL studies that demonstrate impressive performance without syntactic backbones and suggest that syntax information becomes much less important for neural semantic role labeling, especially when paired with recent deep neural network and large-scale pre-trained language models. Despite this notion, the neural SRL field still lacks a systematic and full investigation on the relevance of syntactic information in SRL, for both dependency and both monolingual and multilingual settings.  This paper intends to quantify the importance of syntactic information for neural SRL in the deep learning framework. We introduce three typical SRL frameworks (baselines), sequence-based, tree-based, and graph-based, which are accompanied by two categories of exploiting syntactic information: syntax pruning-based and syntax feature-based. Experiments are conducted on the CoNLL-2005, 2009, and 2012 benchmarks for all languages available, and results show that neural SRL models can still benefit from syntactic information under certain conditions. Furthermore, we show the quantitative significance of syntax to neural SRL models together with a thorough empirical survey using existing models.
\end{abstract}

\section{Introduction}

Semantic role labeling (SRL), namely semantic parsing, is a shallow semantic parsing task that aims to recognize the predicate-argument structure of each predicate in a sentence, such as \textit{who} did \textit{what} to \textit{whom}, \textit{where} and \textit{when}, etc. Specifically, SRL seeks to identify arguments and label their semantic roles given a predicate. SRL is an important method for obtaining semantic information that is beneficial to a wide range of natural language processing (NLP) tasks, including machine translation~\cite{shi-etal-2016-knowledge}, question answering~\cite{berant-etal-2013-semantic,yih-etal-2016-value}, and discourse relation sense classification~\cite{mihaylov-frank-2016-discourse} and relation extraction~\cite{lin-etal-2017-neural}.

\begin{figure}
	\centering
	\includegraphics[scale=1]{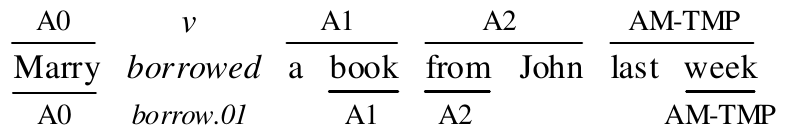}
	\caption{\label{fig:example} Examples of annotations in span (above) and dependency (below) SRL.}
\end{figure}

SRL can be split into four subtasks: predicate detection, predicate disambiguation, argument identification, and argument classification. 
For argument annotation, there are two formulizations (styles). 
One is based on constituents (i.e., phrase or span), while the other is based on dependencies. The other, proposed by the CoNLL-2008 shared task~\cite{surdeanu-etal-2008-conll}, is also called semantic dependency parsing and annotates the heads of arguments rather than phrasal arguments. Figure \ref{fig:example} shows example annotations.

In prior SRL work, considerable attention has been paid to feature engineering, which struggles to capture sufficient discriminative information compared to neural network models, which are capable of extracting features automatically. In particular, syntactic information, including syntactic tree features, has been known to be extremely beneficial to SRL since the large scale of empirical verification of~\citet{punyakanok-etal-2008-importance}. Despite their success, their work suffered from erroneous syntactic input, leading to an unsatisfactory performance.

To alleviate the above issues, \citet{marcheggiani-etal-2017-simple,he-etal-2017-deep} proposed a simple but effective neural model for SRL without syntactic input. Their work suggested that neural SRL does not have to rely on syntactic features, contradicting the belief that syntax is a necessary prerequisite for SRL, which was believed as early as~\citet{gildea-palmer-2002-necessity}. This dramatic contradiction motivated us to make a thorough exploration on syntactic contribution to SRL.

Both span and dependency are effective formal representations for semantics, though it has been unknown which form, span or dependency, would be better for the convenience and effectiveness of semantic machine learning and later applications for a long time. This topic has been roughly discussed in \cite{johansson-nugues-2008-dependency,li2019dependency}, who both concluded that the (best) dependency SRL system at then clearly outperformed the span-based (best) system through gold syntactic structure transformation; however, due to the different requirements of downstream task applications, span and dependency both remain focuses of research. Additionally, the two forms of SRL may benefit from each other joint rather than separated development.  We, therefore, revisit the syntax roles under a more solid empirical basis and explore the syntax roles for the two styles with syntax information in equal quality, respectively.

Recent works on syntax contributions have been limited to individual models and the ways in which syntax has been utilized. The conclusions drawn for syntax roles therefore have some limitations. In order to reduce these limitations, we explored three typical and strong baseline models and two categories of syntactic utilization methods. In addition, pre-trained language models, such as ELMo \cite{peters-etal-2018-deep} and BERT \cite{devlin-etal-2019-bert}, that build contextualized representations, continue to provide gains on NLP benchmarks, and \citet{hewitt-manning-2019-structural} showed that structure of syntax information emerges in the deep models' word representation spaces. Whether neural SRL models can further benefit from explicit syntax information in addition to this implicit syntax information, however, is another issue we consider.


Besides, most of SRL literature is dedicated to impressive performance gains on English, while other multiple languages receive relatively little attention. Although human languages have some basic commonalities in syntactic structure and even different levels of grammar, their differences are also very obvious. The study of syntactic roles needs to be examined in the context of multiple languages for verifying its effectiveness and applicability.

In order to quantitatively evaluate the contribution of syntax to SRL, we adopt the ratios between labeled F$_1$ score for semantic dependencies (Sem-F$_1$) and the labeled attachment score (LAS) for syntactic dependencies, F$_1$ score for syntactic constituents. This ration was first introduced by CoNLL-2008 \cite{surdeanu-etal-2008-conll} Shared Task as an evaluation metric. Considering that various syntactic parsers contribute different syntactic inputs with varying levels of quality, the ratio provides a fairer comparison between syntactically-driven SRL systems, which our empirical study surveys.

\section{Background}

SRL was pioneered by \citet{gildea-jurafsky-2000-automatic}, who used the PropBank conventions \cite{palmer-etal-2005-proposition}.
Conventionally, when identifying predicates, span SRL decomposes to two subtasks: argument identification and argument classification. The former identifies the arguments of a predicate, and the latter assigns them semantic role labels, determining the relations between arguments and predicates. PropBank defines a set of semantic roles for labeling arguments. These roles fall into two categories: \textit{core} and \textit{non-core} roles. The core roles (A0-A5 and AA) indicate different semantics in predicate-argument structure, while the non-core roles are modifiers (AM-\textit{adj}), where \textit{adj} specifies the adjunct type, such as in temporal (AM-TMP) and locative (AM-LOC) adjuncts. For the example shown in Figure \ref{fig:example}, A0 is a proto-agent, representing the \textit{borrower}.

Slightly different from span SRL in argument annotation, dependency SRL labels the syntactic heads of arguments rather than entire phrases, a practice popularized by the CoNLL-2008 and CoNLL-2009 shared tasks\footnote{CoNLL-2008 is an English-only task, while CoNLL-2009 extends to a multilingual one. Their main difference is that predicates have been beforehand indicated for the latter. Or rather, CoNLL-2009 does not need predicate identification, but it is an indispensable subtask for CoNLL-2008.} \cite{surdeanu-etal-2008-conll,hajic-etal-2009-conll}. Furthermore, when no predicate is given, two other indispensable subtasks of dependency SRL are required: predicate identification and predicate disambiguation. The former identifies all predicates in a sentence, and the latter determines the word senses, the specific contextual meanings, of predicates. In the example shown in Figure \ref{fig:example}, \textit{01} indicates the first sense from the PropBank sense repository for predicate \textit{borrowed} in the sentence.

\begin{table*}[!htp]
	\renewcommand\arraystretch{1.3}
	\setlength{\tabcolsep}{5pt}
	\centering
	\caption{A chronicle of related work for span and dependency SRL. SA represents a syntax-aware system (no + indicates syntax-agnostic system). F$_1$ is the result of a single model on the official test set.}
	\scalebox{0.71}{
	\begin{tabular}{llclc|llclc}
		\hline
		
		\hline
		\multicolumn{5}{c|}{\textbf{Span (CoNLL 2005)}} & \multicolumn{5}{c}{\textbf{Dependency (CoNLL 2009)}} \\ 
		\hline
		\textbf{Time} & System  & SA & Method & \textbf{F$_1$} & \textbf{Time} & System & SA & Method & \textit{\textbf{F$_1$}} \\ 
		\hline
		\textbf{2008} & Punyakanok et al. & + &  ILP & \textbf{76.3} & \textbf{2009} & \citeauthor{zhao-etal-2009-multilingual-dependency} & + &   ME & \textit{\textbf{86.2}} \\ 
		\hline
		\textbf{2008} & Toutanova et al. & + & DP & \textbf{79.7} & \textbf{2010} & \citeauthor{bjorkelund-etal-2010-high} & + &  global & \textit{\textbf{86.9}} \\ 
		\hline
		
		\hline
		\multicolumn{1}{l}{\textbf{2015}} & \multicolumn{1}{l}{\textbf{\citeauthor{fitzgerald-etal-2015-semantic}}} &
		\multicolumn{1}{c}{+}  & \multicolumn{1}{l}{structured} & \multicolumn{1}{c}{\textbf{79.4}} & \multicolumn{2}{l}{} & \multicolumn{1}{c}{+} & \multicolumn{1}{l}{structured} & \multicolumn{1}{c}{\textit{\textbf{87.3}}}\\
		\hline
		
		\hline
		\textbf{2015} & \citeauthor{zhou-xu-2015-end} &  & deep BiLSTM & \textbf{82.8} &  &  &  &  &  \\
		\hline
		&  &  &  &  & \textbf{2016} & \citeauthor{roth-lapata-2016-neural} & +  & PathLSTM & \textit{\textbf{87.7}} \\ 
		\hline
		\textbf{2017} & \citeauthor{he-etal-2017-deep} &  & highway BiLSTM & \textbf{83.1} & \textbf{2017} & Marcheggiani et al. &   & BiLSTM  & \textit{\textbf{87.7}} \\ 
		\hline
		&  &  &  &  & \textbf{2017} & \citeauthor{marcheggiani-titov-2017-encoding} & +   & GCNs  & \textit{\textbf{88.0}} \\
		\hline
		\textbf{2018} & \citeauthor{selfatt2018} &  & self-attention & \textbf{84.8} & \textbf{2018} & \citeauthor{he-etal-2018-syntax} (b) & + & ELMo  & \textit{\textbf{89.5}} \\ 
		\hline
		\textbf{2018} & \citeauthor{strubell-etal-2018-linguistically}  & + & self-attention & \textbf{83.9} & \textbf{2018} & \citeauthor{cai-etal-2018-full} &  & biaffine  & \textit{\textbf{89.6}} \\ 
		\hline
		\textbf{2018} & \citeauthor{he-etal-2018-jointly} (a) &  & ELMo  & \textbf{87.4} & \textbf{2018} & \citeauthor{li-etal-2018-unified} (a) & +& ELMo & \textit{\textbf{89.8}} \\ 
		\hline
		
		\hline
		\textbf{2019} & \multicolumn{1}{l}{Li et al. (b) AAAI} &
		\multicolumn{1}{c}{} & \multicolumn{1}{l}{ELMo+biaffine} & \multicolumn{1}{c}{\textbf{87.7}} & \multicolumn{2}{l}{} & \multicolumn{1}{c}{} & \multicolumn{1}{l}{ELMo+biaffine} & \multicolumn{1}{c}{\textit{\textbf{90.4}}}\\
		\hline
		
		\hline
	\end{tabular}
	}
	\label{tab:related work}
\end{table*}

\section{Methodology} 

To fully disclose the predicate-argument structure, typical SRL systems have to perform four subtasks step-by-step or jointly learn and predict the four targets. 
In order to research the role of syntax, we evaluate our systems in two separate settings: being given the predicate and not being given the predicate. For the first setting, our backbone models all only focus on the identification and labeling of arguments. We use the pre-identified predicate information when the predicate is provided in the corpus and adopt a sequence tagging model to perform predicate disambiguation. In the second condition, we do the work of predicate identification and disambiguation in one sequence tagging model. In summary, we focus on three backbone models for argument identification and disambiguation and feed the predicates into the models as features.

\subsection{Factorization and Modeling}

\begin{figure}
	\centering
	\includegraphics[width=1.0\textwidth]{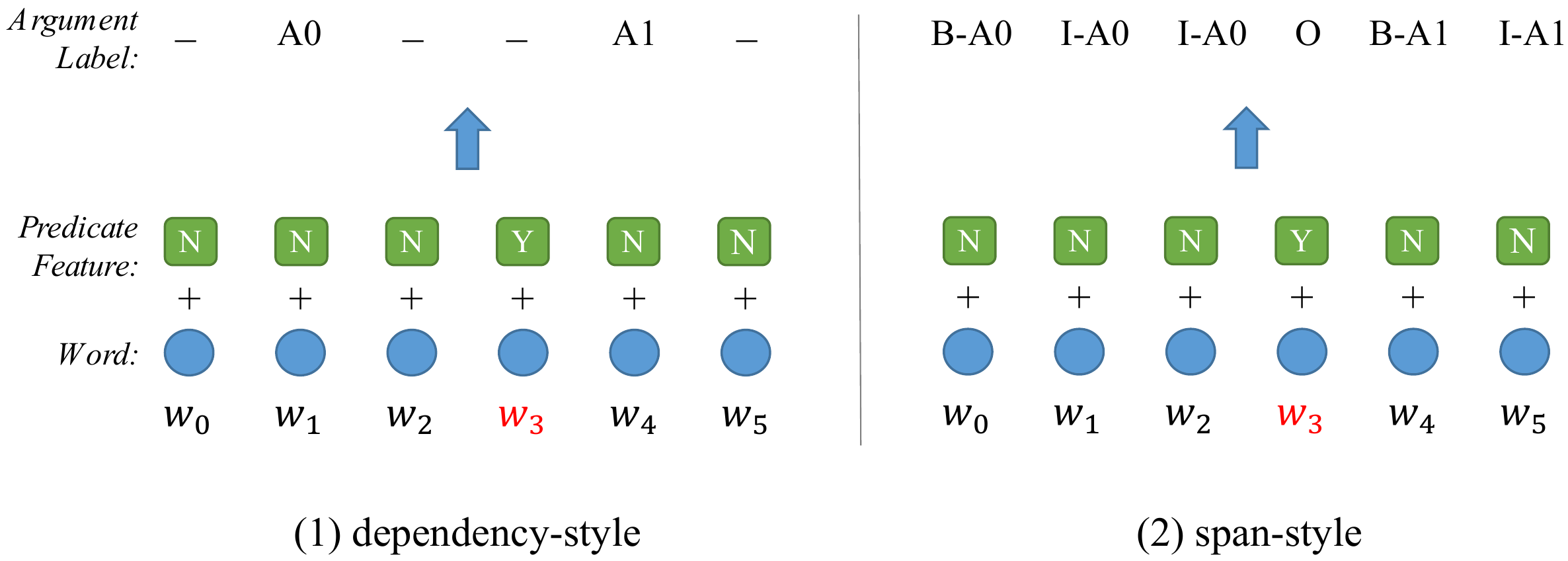}
	\caption{An example of sequence-based factorization.} \label{fig:factorization_st}
\end{figure}

\begin{figure}
	\centering
	\includegraphics[width=1.0\textwidth]{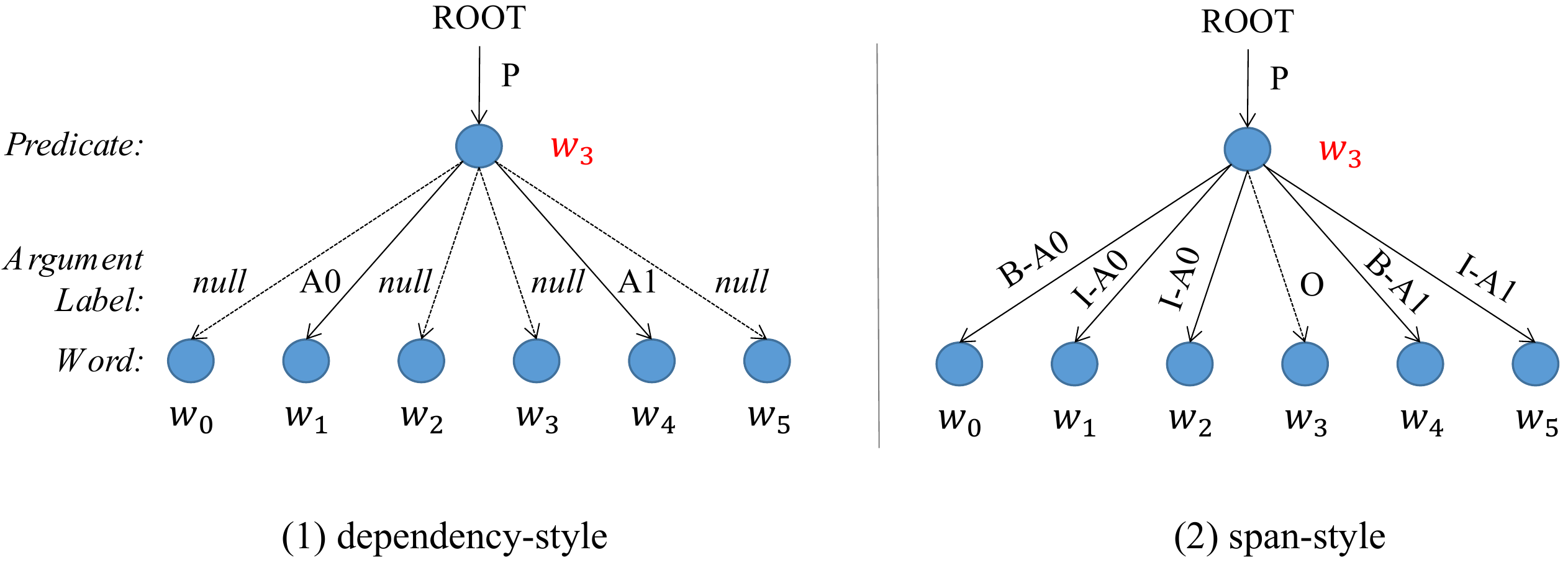}
	\caption{An example of tree-based factorization.} \label{fig:factorization_tree}
\end{figure}

We summarize and present three typical baseline models, which are based on methods of factorizing and modeling for semantic graphs in SRL:
\begin{itemize}
	\item \textbf{Sequence-based:} As shown in Figure \ref{fig:factorization_st}, the semantic dependency graph of SRL is decomposed by predicates. The arguments for each predicate consist of a sequence, either under dependency or span-style. Notably, an extra Begin-Inside-Outside (BIO) conversion step is required for span-style argument labels. This decomposition is very simple and efficient. In the baseline model of this factorization, the predicate needs to be input as a source feature, which allows the model to produce different inputs for different target argument sequences. Predicate-specific embeddings are usually used for this reason.
	\item \textbf{Tree-based:} Embedding differentiation by relying on the predicate indicate inputs is only a soft constraint and prompt. This feature may be lost due to forgetting mechanisms such as dropout in the encoder, which may limit SRL model performance. The tree-based method also decomposes the semantic dependency graph to trees with a depth of 2 according to the predicate. The predicate is the child node of $\text{ROOT}$, and all others are child nodes of the predicate, as shown in Figure \ref{fig:factorization_tree}. An empty relation $null$ is set between the non-arguments and the predicate in order to fill the tree. The tree-based factorization can be thought as an enhanced version of the sequence-based factorization, as the predicate is more prominent and obvious to specify. To emphasize a given predicate being handled, predicate-specific embeddings are also applied.
	\item \textbf{Graph-based:} Sequence-based and tree-based models score the $<argument, label>$ tuple structure for a determined predicate. The graph-based method further extends this mode; specifically it adjusts for the case of undetermined predicates and models on the semantic dependency graph directly to output a $<predicate, argument, label>$ triple structure, allowing the model handle labeling multiple predicates and arguments at the same time. This mode not only handles instances without given predicates but also allows instances with given predicates to be enhanced by predicate-specific embeddings. Using the dependency-style, the graph-based method is a trivial extension of the tree-based method. The span-style situation is not so simple because of its argument structure. To account for this, graph-based models enumerate and sort all possible spans, takes them as candidate arguments, and scores them with the candidate sets of predicates.
\end{itemize}

The above methods cover most mainstream SRL models to our best knowledge. In the sequence-based and tree-based methods, the BIO conversion when using the span style, and some works use Conditional Random Fields (CRFs) to model this constraint. 

\begin{figure}
	\centering
	\includegraphics[width=1.0\textwidth]{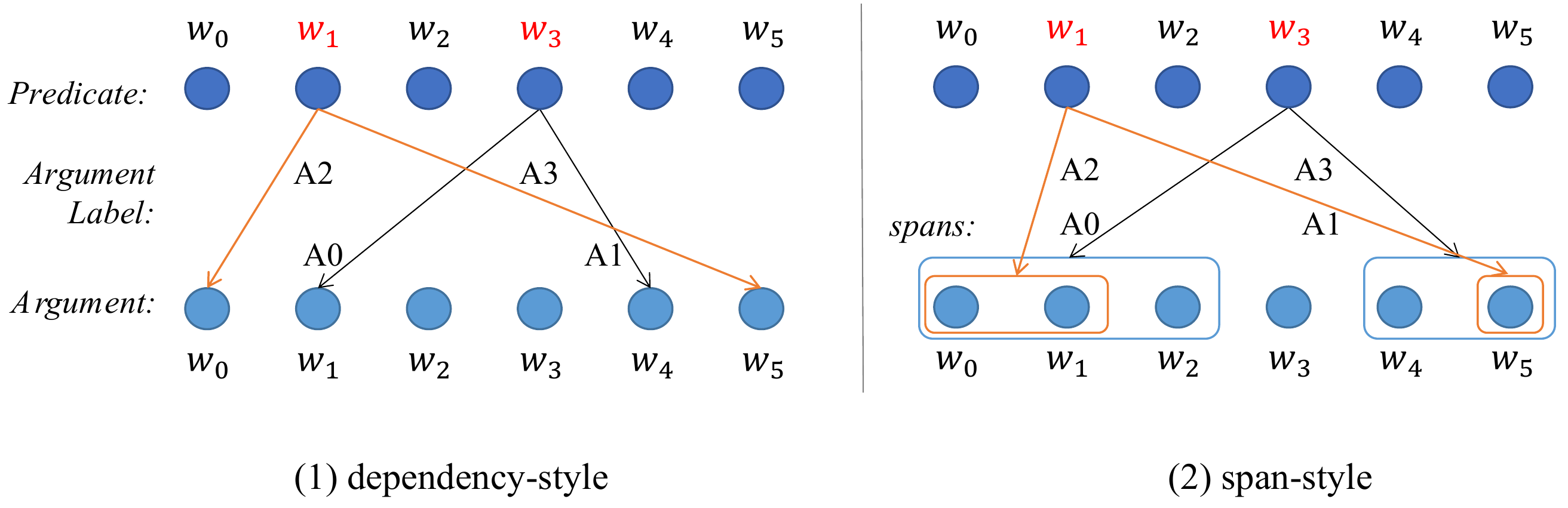}
	\caption{An example of graph-based factorization. We omit the dashed line between non-predicates and non-arguments, i.e., the empty relation $null$, here.} \label{fig:factorization_graph}
\end{figure}

\subsection{Baseline Implementation}

This subsection presents the basic neural SRL models separately under the three previous aforementioned methods. In order to make fair comparisons, we make the architectures of these models as similar as possible.

\paragraph{Word Representation}\label{sec:word}

\begin{figure}
	\centering
	\includegraphics[scale=0.73]{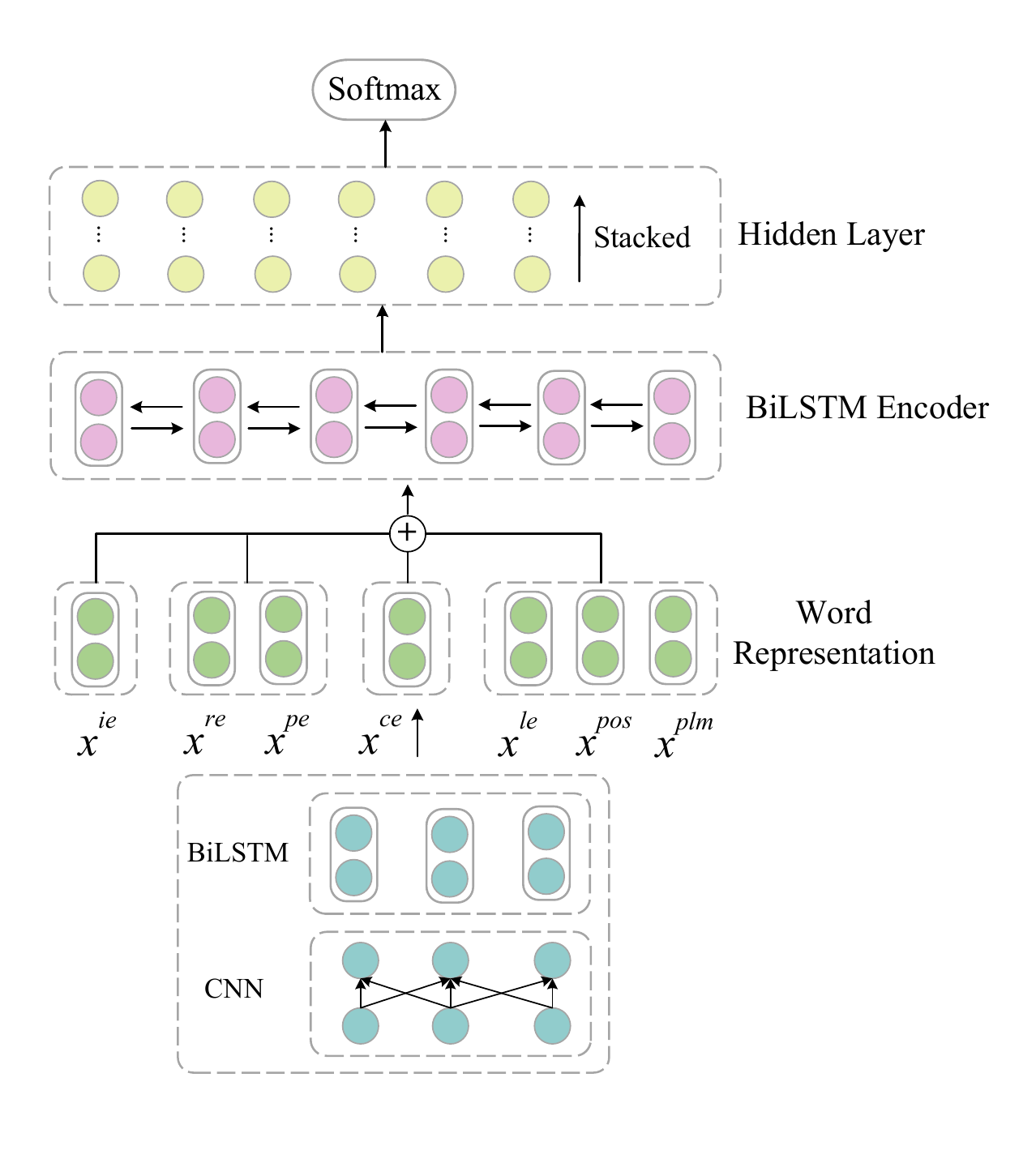}
	\caption{The sequence-based argument labeling baseline model.} \label{fig:st-model}
\end{figure}

We produce a predicate-specific word representation $x_i$ for each word $w_i$ in the sequence $w = \{w_1, \cdots, w_n\}$, where $i$ stands for the word position in an input sequence, and $n$ is the length of this sequence, following~\citet{marcheggiani-etal-2017-simple}. 
In this work, word representation $\bm{e}_i$ is the concatenation of four types of features: a predicate-specific feature and, character-level, word-level and linguistic features. Since previous works demonstrated that the predicate-specific feature is helpful in promoting the role labeling process, we leverage a predicate-specific indicator embedding $\bm{e}_i^{ie}$ to indicate whether a word is a predicate when predicting and labeling the arguments for each given predicate.
At the character level, we exploit a convolutional neural network (CNN) with a bidirectional LSTM (BiLSTM) to learn character embedding $\bm{e}_i^{ce}$. As shown in Figure \ref{fig:st-model}, the representation calculated by the CNN is fed as input to the BiLSTM. At the word level, we use a randomly initialized word embedding $\bm{e}_i^{re}$ and a pre-trained word embedding $\bm{e}_i^{pe}$. For linguistic features, we employ a randomly initialized lemma embedding $\bm{e}_i^{le}$ and a randomly initialized POS tag embedding $\bm{e}_i^{pos}$. 
To further enhance the word representation, we leverage an optimal external representation $\bm{e}^{plm}_i$ from pre-trained language models. 
The resulting word representation is concatenated as $\bm{e}_i = [\bm{e}_i^{ie}, \bm{e}_i^{ce}, \bm{e}_i^{re}, \bm{e}_i^{pe}, \bm{e}_i^{le}, \bm{e}_i^{pos}, \bm{e}_i^{plm}]$.

\paragraph{Sequence Encoder}\label{sec:encoder}

As Long short-term memory (LSTM) networks~\cite{hochreiter1997long} have shown significant representational effectiveness to NLP tasks, we thus use BiLSTM as the sentence encorder. Given a sequence of word representation $x=\{\bm{e}_1, \bm{e}_2, \cdots, \bm{e}_n\}$ as input, the $i$-th hidden state $\bm{g}_i$ is encoded as follows:
\begin{equation*}
\bm{g}^f_i = LSTM^\mathcal{F}\left(\bm{e}_i, \bm{g}^f_{i-1}\right),\ \ \  
\bm{g}^b_i = LSTM^\mathcal{B}\left(\bm{e}_i, \bm{g}^b_{i+1}\right),\ \ \  
\bm{g}_i  = \bm{g}^f_i \oplus \bm{g}^b_i, 
\end{equation*}
where $LSTM^\mathcal{F}$ denotes the forward LSTM transformation and $LSTM^\mathcal{B}$ denotes the backward LSTM transformation. $\bm{g}^f_i$ and $\bm{g}^b_i$ are the hidden state vectors of the forward LSTM and backward LSTM, respectively.

\paragraph{Scorer in Sequence-based Model}

In the sequence-based models, i.e., the sequence tagging model, to get the final predicted semantic roles, stacked multi-layer perceptron (MLP) layers on the top of BiLSTM networks are usually exploited, which take as input the hidden representation $h_i$ of all time steps and employ \textit{ReLU} activations between the hidden layers. Finally, A softmax layer is used over the outputs to maximize the likelihood of labels.

\paragraph{Scorer in Tree-based Model}

As in the sequence-based model, to predict and label arguments for a given predicate, a role classifier is employed on top of the BiLSTM encoder. Some works like \cite{marcheggiani-etal-2017-simple} show that incorporating the predicate's hidden state in their role classifier enhances the model performance, while we argue that a more natural way to incorporate the syntactic information carried by the predicate is to employ the attentional mechanism. We adopt the recently introduced biaffine attention \cite{dozat2017deep} to enhance our role scorer. Biaffine attention is a natural extension of bilinear attention \cite{luong-etal-2015-effective}, which is widely used in neural machine translation (NMT). 

\textbf{Nonlinear Affine Transformation} 
Usually, a BiLSTM decoder takes the concatenation $\bm{g}_i$ of the hidden state vectors as output for each hidden state; however, in the SRL context, the encoder is supposed to distinguish the currently considered predicate from its candidate arguments. As noted in \cite{dozat2017deep}, applying a multi-layer perceptron (MLP) to the recurrent output states before the classifier has the advantage of stripping away irrelevant information for the current decision.  Therefore, to distinguish the currently considered predicate from its candidate arguments in an SRL context, we perform two distinct affine transformations with a nonlinear activation on the hidden state $\bm{g}_i$, mapping it to vectors with smaller dimensionality:
\begin{equation*}
\bm{h}_i^{(pred)} = ReLU \left(\bm{W}^{(pred)}\bm{g}_i + \bm{b}^{(pred)} \right),\ \ 
\bm{h}_i^{(arg)} = ReLU \left(\bm{W}^{(arg)}\bm{g}_i + \bm{b}^{(arg)} \right),
\end{equation*}
where $ReLU$ is the rectilinear activation function \cite{nair2010rectified}, $\bm{h}_i^{(pred)}$ is the hidden representation for the predicate and $\bm{h}_i^{(arg)}$ is the hidden representation for the candidate arguments. 

By performing such transformations over the encoder output to feed the scorer, the scorer may benefit from deeper feature extraction. This leads to two benefits. First, instead of keeping both features learned by the two distinct LSTMs, the scorer ideally is now able to learn features composed from both recurrent states together with reduced dimensionality. Second, it provides the ability to map the predicates and the arguments into two distinct vector spaces, which is essential for our tasks, since some words can be labeled as predicates and arguments simultaneously. Mapping a word into two different vectors can help the model disambiguate its role in different contexts. 

\textbf{Biaffine Scoring} 
In the standard NMT context, given a target recurrent output vector $h_i^{(t)}$ and a source recurrent output vector $h_j^{(s)}$, a bilinear transformation calculates a score $s_{ij}$ for the alignment:
\begin{equation*}
s_{ij} = \bm{h}_i^{\top (t)} \bm{W} \bm{h}_j^{(s)},
\end{equation*}

However, in a traditional classification task, the distribution of classes is often uneven, and the output layer of the model normally includes a bias term designed to capture the prior probability $P(y_i = c)$ of each class, with the rest of the model focusing on learning the likelihood of each class given the data $P (y_i = c|x_i )$. \citet{dozat2017deep} incorporated the bias terms into the bilinear attention to address this uneven problem, resulting in a biaffine transformation. The biaffine transformation is a natural extension of the bilinear transformation and the affine transformation. In the SRL task, the distribution of the role labels is similarly uneven, and the problem worsens after introducing the additional $\text{ROOT}$ node and $null$ label; directly applying the primitive form of bilinear attention would fail to capture the prior probability $P(y_i=c_k)$ for each class. Thus, introducing the biaffine attention in our model would be extremely helpful for semantic role prediction.

It is worth noting that in our model, the scorer aims to assign a score for each specific semantic role. Besides learning the prior distribution for each label, we wish to further capture the preferences for the label that a specific predicate-argument pair can take. Thus, our biaffine attention contains two distinct bias terms: 
\begin{align}
\bm{s}_{i,j} = \textit{Biaffine}(\bm{h}_i^{(pred)}, \bm{h}_i^{(arg)}) = \ & \bm{h}_i^{\top (arg)} \bm{W}^{(role)} \bm{h}_j^{(pred)}\label{bilinear}\\
& + \bm{U}^{(role)} \bm{h}_i^{(arg)}\oplus \bm{h}_j^{(pred)} \label{pair_bias}\\
& + \bm{b}^{(role)} \label{class_bias},
\end{align}
where $\bm{W}^{(role)}$, $\bm{U}^{(role)}$ and $\bm{b}^{(role)}$ are parameters that will be updated by some gradient descent methods in the learning process. There are several points that should be paid attention to in the above biaffine transformation. First, since our goal is to predict the label for each pair of $\bm{h}_i^{(arg)}$, $\bm{h}_j^{(pred)}$, the output of our biaffine transformation should be a vector of dimensionality $N_r$ instead of a real value, where $N_r$ is the number of all the candidate semantic labels. Thus, the bilinear transformation in Eq. (\ref{bilinear}) maps two input vectors into another vector. This can be accomplished by setting $\bm{W}^{(role)}$ as a $(d_h \times N_r \times d_h)$ matrix, where $d_h$ is the dimensionality of the hidden state vector. Similarly, the output of the linear transformation in Eq. (\ref{pair_bias}) is also a vector by setting $\bm{U}^{(role)}$ as a $(N_r \times 2 d_h)$ matrix. Second, Eq. (\ref{pair_bias}) captures the preference of each role (or sense) label and is conditioned on taking the $j$-th word as a predicate and the $i$-th word as an argument. Third, the last term $\bm{b}^{(role)}$ captures the prior probability of each class $P(y_i = c_k)$. Notice that Eq. (\ref{pair_bias}) and (\ref{class_bias}) capture different kinds of bias for the latent distribution of the label set.

Given a sentence of length $n$, for one of its predicates $w_j$, the scorer outputs a score vector $\{\bm{s}_{1,j}, \bm{s}_{2,j}, \cdots, \bm{s}_{n,j}\}$. Then, our model picks as its output the label with the highest score from each score vector: $y_{i,j}=\mathop{\arg\max}_{1\leq k\leq N_r} (\bm{s}_{i,j}[k])$, where $\bm{s}_{i,j}[k]$ denotes the score of the $k$-th candidate in the semantic label vocabulary with size $N_r$.

\begin{figure}
	\centering
	\includegraphics[width=1.0\textwidth]{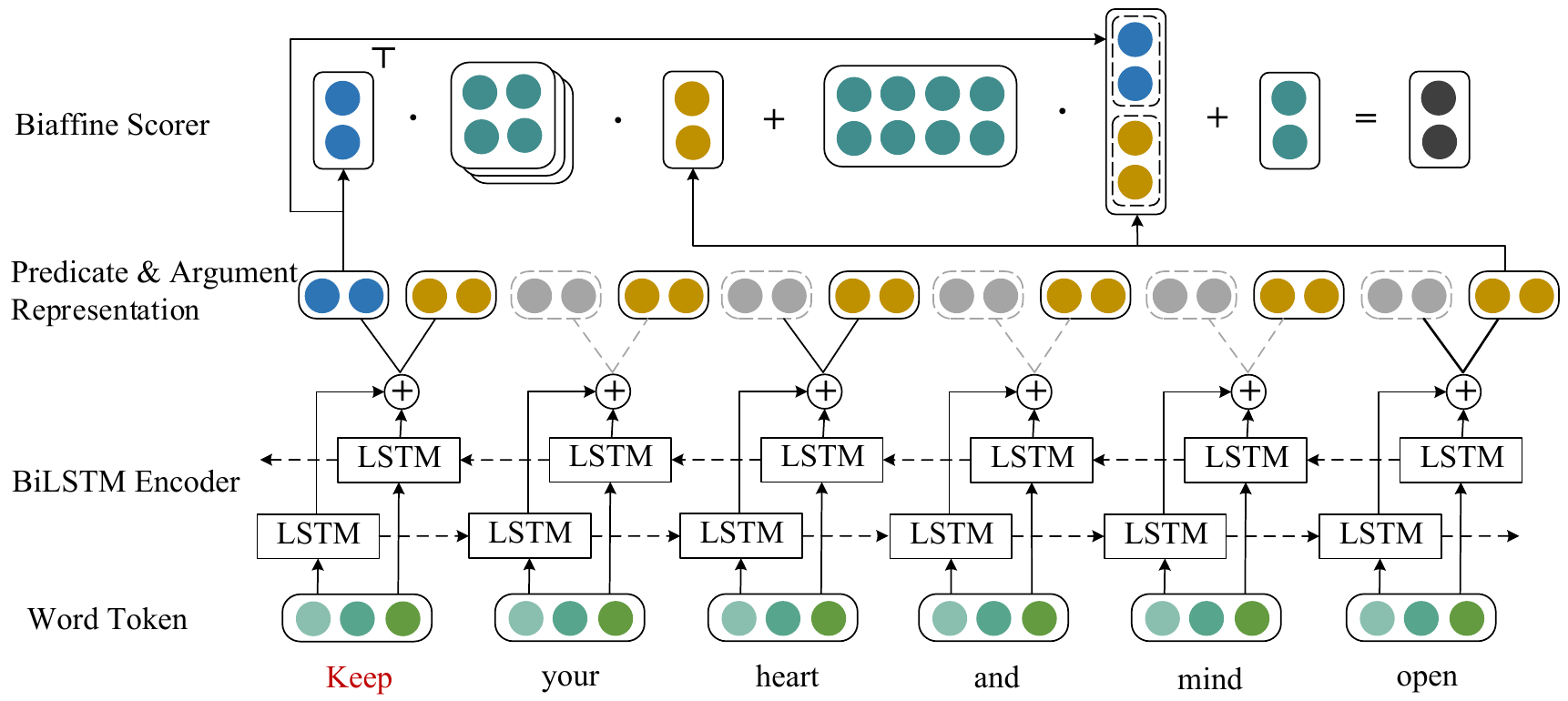}
	\caption{The tree-based argument labeling baseline model.}\label{fig:tree-model} 
\end{figure}

\paragraph{Scorer in Graph-based Model} 
As in the scorer of the tree-based model, the graph-based model also uses the biaffine scorer to score the predicate-argument structure. 
Similarly, we also employ a nonlinear affine transformation on the top of the BiLSTM encoder. In the sequence-based and tree-based models, dependency and span-style arguments are converted into a consistent label sequence, while the graph-based model treats arguments as independent graph nodes. In order to unify the two styles of models, we introduce a unified argument representation that can handle both styles of SRL tasks.

In the sentence $w_1, w_2, \dots, w_n$, the model aims to predict a set of predicate-argument-relation tuples $\mathcal{Y} \in \mathcal{P} \times \mathcal{A} \times \mathcal{R}$, where $\mathcal{P} = \{w_1, w_2, ..., w_n\}$ is the set of all possible predicate tokens, $\mathcal{A} = \{(w_i,\dots,w_j) | 1 \leq i \leq j \leq n \}$ includes all the candidate argument spans or dependencies\footnote{When $i$=$j$, span reduces to dependency.}, and $\mathcal{R}$ is the set of the semantic roles.
For dependency SRL, we assume single word argument spans and thus limit the length of candidate argument to be 1, so our model uses the $h^{arg}$ as the final argument representation $\bm{h}^{(arg)'}$ directly. For span SRL, we utilize the span representation from \cite{lee-etal-2017-end}. Each candidate span representation $\bm{h}^{(arg)'}$ is built by
\begin{equation}
\bm{h}^{(arg)'}= [\bm{h}^{(arg)}_{\textit{START}} , \bm{h}^{(arg)}_{\textit{END}} , \bm{h}_\lambda , size(\lambda)],\nonumber
\end{equation}
where $\bm{h}_{START}^{arg}$  and $\bm{h}_{END}^{arg}$ are boundary representations, $\lambda$ indicates a span, $size(\lambda)$ is a feature vector encoding the size of span, and $h_\lambda$ is the specific notion of headedness learned by the attention mechanism \cite{bahdanau2014neural} over words in each span (where $t$ is the position inside span) as follows:
$$\mu^a_t = \textbf{w}_{attn} \cdot \textbf{MLP}_{attn} (\bm{h}^{(arg)}_t),  \quad \nu_t = \frac{\exp(\mu^a_t)}{\sum_{k=\textit{START}}^{\textit{END}} \exp(\mu^a_k)},$$ 
$$h_{\lambda} = \sum_{t=\textit{START}}^{\textit{END}} \nu_t \cdot \bm{h}^{(arg)}_t.$$

\textbf{Candidate Pruning}  The number of candidate arguments for a sentence of length $l$ is $\textit{O}(l^2)$ for span SRL and $\textit{O}(l)$ for dependency. As the model deals with $\textit{O}(l)$ possible predicates, the computational complexity is $\textit{O}(l^3 \cdot |\mathcal{R}|)$ for span and $\textit{O}(l^2 \cdot |\mathcal{R}|)$  for dependency, both of which, are too computationally expensive.

To address this issue, we attempt to prune candidates using two beams for storing the candidate arguments and predicates with size $\beta_pn$ and $\beta_an$ where $\beta_p$ and $\beta_a$ are two manually set thresholds, a method inspired by \cite{he-etal-2018-jointly}. First, the predicate and argument candidates are ranked according to their predicted scores ($\phi_p$ and $\phi_a$, respectively), and then we reduce the predicate and argument candidates with defined beams. Finally, we take the candidates from the beams for use in label prediction. Such pruning will reduce the overall number of candidate tuples to  $\textit{O}(n^2 \cdot |\mathcal{R}|)$ for both types of tasks. Furthermore, for span SRL, we set the maximum length of candidate arguments to $\mathcal{L}$, which may decrease the number of candidate arguments to $\textit{O}(n)$. Specifically, for predicates and arguments, we introduce two unary scores based on their candidates for ranking:
$$\phi_p = \textbf{w}_p \textbf{MLP}^s_p(g^p), \quad \phi_a = \textbf{w}_a \textbf{MLP}^s_a(g_f^a).$$

After pruning, we also adopt the biaffine scorer as in the tree-based models:
\begin{align}
\Phi_{r}(p, a) &= \textit{Biaffine}(\bm{h}^{(pred)}, \bm{h}^{(arg)'}).
\end{align}

\begin{figure}
	\centering
	\includegraphics[width=1.0\textwidth]{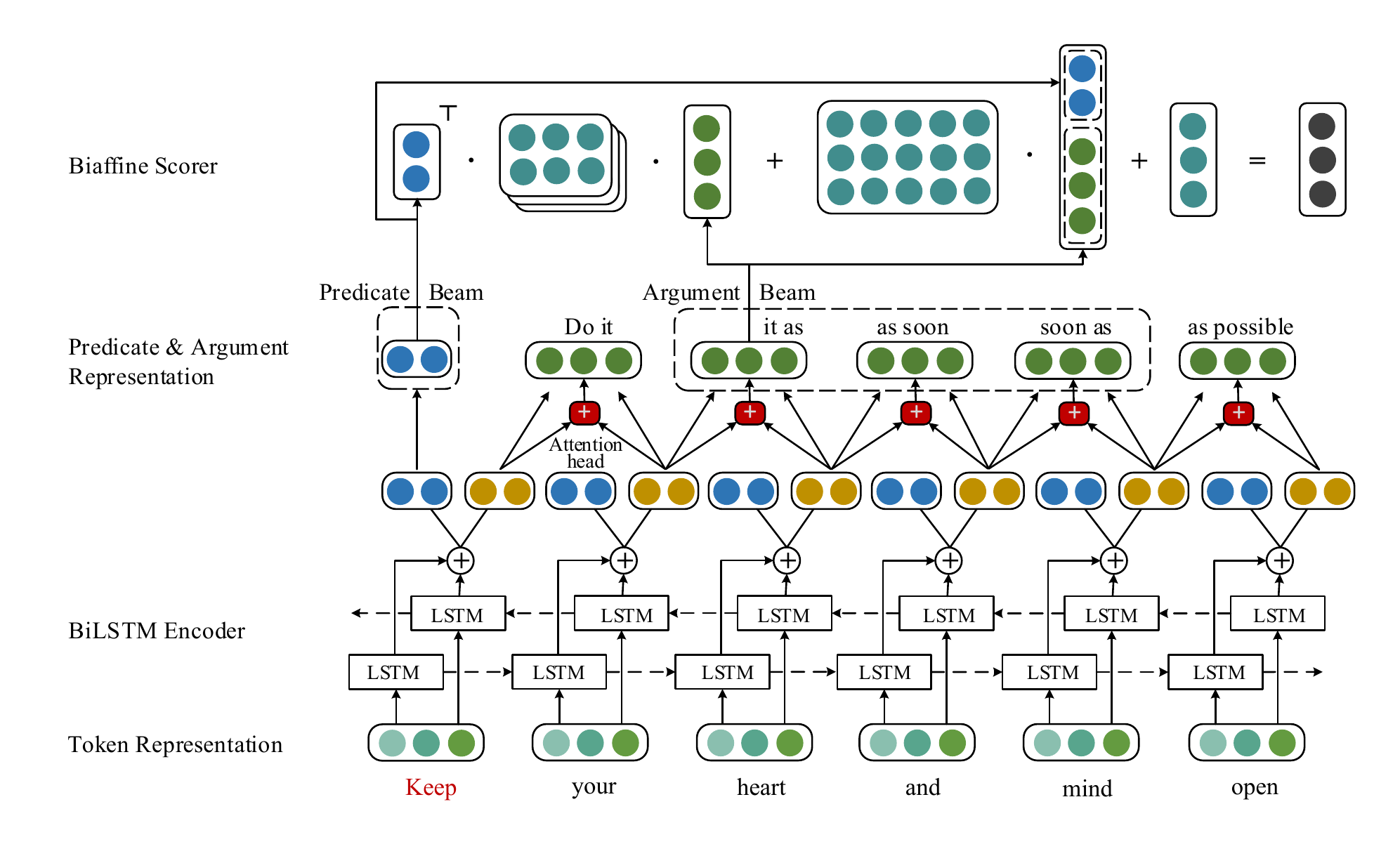}
	\caption{The graph-based argument labeling baseline model.}\label{fig:graph-model} 
\end{figure}

\section{Syntax Utilization}

In this section, we present two types of syntax utilization: syntax-based argument pruning and syntax feature integration.

\subsection{Syntax-based Argument Pruning}

\paragraph{Hard Pruning} The argument structure for each known predicate will be discovered by our argument labeler using the possible arguments (candidates) set. Most SRL works in the pre-NN era select words according to a syntactic parse tree based around the predicate to prune the sequence, a strategy we refer to as hard pruning.
In the NN model, we can also borrow this hard pruning strategy to enhance the SRL baseline, and it is one way of using syntax information. Specifically, before inputting to the model, we use the argument pruning algorithm to get a filtered sequence $w_f = \{w_1,\dots,w_{f}\}$ for each predicate. Then, we replace the original sequence with this one and input it to the SRL model. 

As pointed out by~\citet{punyakanok-etal-2008-importance}, syntactic information is most relevant in identifying the arguments, and the most crucial contribution of full parsing is in the pruning stage. In this paper, we propose a $k$-order argument hard pruning algorithm inspired by~\citet{zhao-etal-2009-semantic}. First, for node $n$ and its descendant $n_d$ in a syntactic dependency tree, we define the $order$ to be the \textit{distance} between the two nodes, denoted as $\mathcal{D}(n,n_d)$. Then, we define $k$-order descendants of $n$ as descendants that satisfy $\mathcal{D}(n,n_d)=k$ and a $k$-order traversal that visits each node from the given node to its descendant nodes within $k$-th order. Note that the definition of $k$-order traversal is somewhat different from a traditional tree traversal in terminology.

\begin{figure}
	\centering
	\includegraphics[scale=1.1]{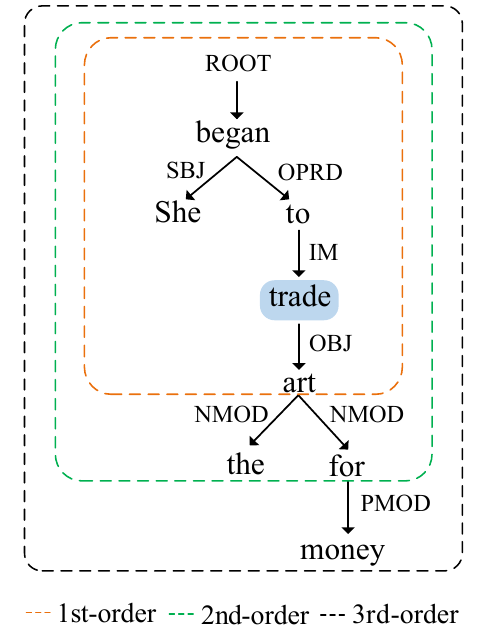}
	\caption{An example of \textit{first}-order, \textit{second}-order and \textit{third}-order argument pruning. The shaded part indicates the given predicate.} \label{fig:synpath}
\end{figure}

A brief description of the proposed $k$-order pruning algorithm is given as follows: initially, we set a given predicate as the current node in a syntactic dependency tree. Then, we collect all its argument candidates using a $k$-order traversal. Afterward, we reset the current node to its syntactic head and repeat the previous step until we reach the root of the tree. Finally, we collect the root and stop. The $k$-order argument algorithm is presented in Algorithm \ref{alg:k-ord} in detail. An example of a syntactic dependency tree for the sentence \textit{She began to trade the art for money} is shown in Figure \ref{fig:synpath}.

\begin{algorithm}[t] 
	\caption{The $k$-order argument pruning algorithm.} 
	\label{alg:k-ord} 
	\begin{algorithmic}[1]
		\renewcommand{\algorithmicrequire}{\textbf{Input:}}
		\renewcommand{\algorithmicensure}{\textbf{Output:}}
		\REQUIRE A predicate $p$, the root node $r$ given a syntactic dependency tree $T$, the order $k$
		\ENSURE The set of argument candidates $S$
		\STATE \textbf{initialization} set $p$ as current node $c$, $c=p$
		\FOR{each descendant $n_i$ of $c$ in $T$}
		\IF{$\mathcal{D}(c,n_i)\leq k$ and $n_i \notin S$}
		\STATE $S=S+n_i$
		\ENDIF
		\ENDFOR
		\STATE find the syntactic head $c_h$ of $c$, and let $c=c_h$
		\IF{$c=r$}
		\STATE  $S=S+r$
		\ELSE
		\STATE \textbf{goto} step 2 
		\ENDIF
		\RETURN argument candidates set $S$
	\end{algorithmic} 
\end{algorithm}

The main reasons for applying the extended $k$-order argument pruning algorithm are two-fold. First, previous standard pruning algorithms may impede the argument coverage too much, even though arguments do usually tend to surround their predicates at a close distance. For a sequence tagging model that has been applied, the algorithm can effectively handle the imbalanced distribution between arguments and non-arguments, which would be poorly handled by early argument classification models that commonly adopt the standard pruning algorithm. Second, the extended pruning algorithm provides a better trade-off between computational cost and performance by carefully tuning $k$.

\paragraph{Soft Pruning} For word pair classification modeling, one major performance bottleneck is caused by unbalanced data. This is especially pertinent for SRL, where more than 90\% of argument candidates are non-arguments. The syntax-based hard pruning methods are thus proposed to alleviate the imbalanced distribution; however, these do not extend well to other baselines and languages and even hinder syntax-agnostic SRL models as \cite{cai-etal-2018-full} demonstrated using different $k$ values on English. This hindrance might result because this pruning method breaks up the whole sentence, leading the BiLSTM encoder to take the incomplete sentence as input and fail to learn sentence representation sufficiently.

To alleviate such a drawback from the previous syntax-based pruning methods, we propose a novel pruning rule extraction method based on syntactic parse trees that generally suits diverse baselines at the same time. In detail, we add an argument pruning layer guided by syntactic rules following BiLSTM layers, which can absorb the syntactic clues simply and effectively. 

\textbf{Syntactic Rule} Considering that all arguments are specific to a particular predicate, it has been observed that the distances between predicates and their arguments on syntactic trees are generally within a certain range for most languages. Therefore, we introduce a language-specific rule based on syntactic dependency parses to prune some unlikely arguments. We call this rule the syntactic rule. Specifically, given a predicate $p$ and its argument $a$, we define $d_p$ and $d_a$ to be the distance from $p$ and $a$ to their nearest common ancestor node (namely, the root of the minimal subtree that includes $p$ and $a$), respectively. For example, $0$ denotes that a predicate or argument itself is their nearest common ancestor, while $1$ represents that their nearest common ancestor is the parent of predicate or argument. Then, we use the distance tuple ($d_p$, $d_a$) as their relative position representation inside the parse tree. Finally, we make a list of all tuples ordered according to how many times that each distance tuple occurs in the training data, which is counted for each language independently. 

It is worth noting that our syntactic rule is determined by the top-$k$ frequent distance tuples. During training and inference, the syntactic rule takes effect by excluding all candidate arguments whose predicate-argument relative positions in the parse tree are not in the list of top-$k$ frequent tuples.

\begin{figure}
	\centering
	\includegraphics[scale=1.2]{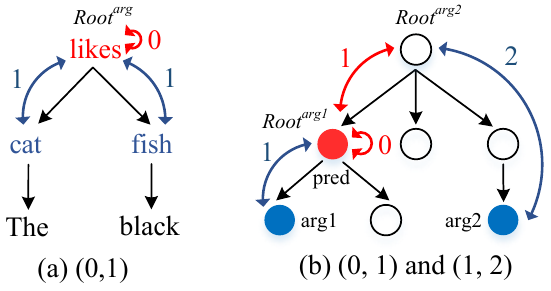}
	\caption{Syntactic parse tree examples (dependency relations are omitted). Red represents the current predicate, and blue indicates its arguments.}\label{fig:multi-syntax} 
\end{figure}

Figure \ref{fig:multi-syntax} shows simplified examples of a syntactic dependency tree. Given an English sentence in Figure \ref{fig:multi-syntax}(a), the current predicate is \textit{likes}, whose arguments are \textit{cat} and \textit{fish}. For \textit{likes} and \textit{cat}, the predicate (\textit{likes}) is their common ancestor (denoted as $Root^{arg}$) according to the syntax tree. Therefore, the relative position representation of the predicate and argument is $(0,1)$, and it is the same for \textit{likes} and \textit{fish}. As for the right side in Figure \ref{fig:multi-syntax}, suppose the marked predicate has two arguments$-$$arg1$ and $arg2$, the common ancestors of the predicate and arguments are respectively $Root^{arg1}$ and $Root^{arg2}$. In this case, the relative position representations are $(0,1)$ and $(1,2)$.  

\textbf{Argument Pruning Method}  To maintain the sequential inputs through the whole sentence, we propose a novel syntax-based method to softly prune arguments, which is unlike most existing works \cite{xue-palmer-2004-calibrating,zhao-etal-2009-multilingual-dependency,he-etal-2018-syntax} with hard pruning strategies that prune argument candidates in the pre-processing stage. The soft pruning strategy is very straightforward. In the argument pruning layer, our model drops these candidate arguments (more exactly, BiLSTM representations) that do not comply with the syntactic rule. In other words, only the predicates and arguments that satisfy the syntactic rule will be output to next layer.

\paragraph{Constituent Pruning} 
In dependency SRL, argument candidates are pruned by a heuristic search over the dependency syntax tree; however, correspondences like those between dependency arcs and dependency semantic relations are infeasible to obtain in constituent syntax trees and span-based SRL. Therefore, we adopt a new constituent-based argument pruning method.

Constituency syntax breaks a sentence into constituents (i.e., phrases or spans), which naturally forms a constituency tree in a top-down fashion. In contrast with the dependency syntax tree, words can only be the terminals in a constituency tree, while the non-terminals are phrases with types. In span SRL, each argument corresponds to a constituent in a constituency tree, which can thus be used to generate span argument candidates given the predicates \cite{xue-palmer-2004-calibrating,carreras-marquez-2005-introduction}. \citet{punyakanok2005necessity} showed that constituency trees offer high-quality argument boundaries.
	
Considering that span SRL models only occasionally violate the syntactic constraints (some candidate arguments may not be constituents), we attempt to prune unlikely arguments based on these constraints, essentially ruling out the likely impossible candidates, albeit at the cost of missing some of the rare violating arguments. 

In order to utilize such constituent boundaries in the constituency tree and help decide argument candidates, we extract all boundaries for a constituent $c$ to form a set $boundaryset = \{(\text{S}\text{\tiny{TART}}(c),\text{E}\text{\tiny{ND}}(c))\}$. We also define an argument pruning layer that drops candidate arguments whose boundaries are not in this set. It is worth noting that since span arguments are converted to BIO labels under the sequence-based and tree-based modeling approaches of span SRL, there is no explicit correspondence between the existing arguments and the constituents, so constituent-based argument pruning is not applicable to the sequence-based and tree-based modeling approaches. We only consider this syntax enhancement when using graph-based modeling.

\subsection{Syntax Feature Integration}

In addition to guiding argument pruning, another major use of syntax information is serving as a syntax-aware feature in addition to the contextualized representation, thereby enhancing the argument labeler. To integrate the syntactic information into sequential neural networks, we employ a syntactic encoder on top of the BiLSTM encoder. 

Specifically, given a syntactic dependency tree $T$, for each node $n_k$ in $T$, let $C(k)$ denote the syntactic children set of $n_k$, $H(k)$ denote the syntactic head of $n_k$, and $L(k, \cdot)$ be the dependency relation between node $n_k$ and those that have a direct arc from or to $n_k$. Then, we formulate the syntactic encoder as a transformation $f^{\tau}$ over the node $n_k$, which may take some of $C(k)$, $H(k)$, or $L(k, \cdot)$ as input and compute a syntactic representation $v_k$ for node $n_k$, namely, $v_k = f^{\tau}(C(k), H(k), L(k, \cdot), x_k)$. 
When not otherwise specified, $x_k$ denotes the input feature representation of $n_k$, which may be either the word representation $e_k$ or the output of BiLSTM $h_k$. $\sigma$ denotes the logistic sigmoid function, and $\odot$ denotes the element-wise multiplication.

In practice, the transformation $f^{\tau}$ can be any syntax encoding method. In this paper, we will consider three types of syntactic encoders: syntactic graph convolutional network (Syntactic GCN), syntax aware LSTM (SA-LSTM), and tree-structured LSTM (Tree-LSTM). 

\paragraph{Syntactic GCN}
The GCN \cite{gcn2017} was proposed to induce the representations of nodes in a graph based on the properties of their neighbors. Given its effectiveness, \citet{marcheggiani-titov-2017-encoding} introduced a generalized version for the SRL task, namely syntactic GCN, and showed that the syntactic GCN is effective in incorporating syntactic information into neural models.

The syntactic GCN captures syntactic information flowing in two directions: one from heads to dependents (along), and the other from dependents to heads (opposite). Besides, it also models the information flows from a node to itself; namely, it assumes that a syntactic graph contains a self-loop for each node. Thus, the syntactic GCN transformation of a node $n_k$ is defined on its neighborhood $N(k) = C(k) \cup H(k) \cup \{n_k\}$. For each edge that connects $n_k$ and its neighbor $n_j$, we can compute a vector representation, 
$$u_{k, j} = W^{dir(k, j)} x_j + b^{L(k, j)},$$
where $dir(k, j)$ denotes the direction type (along, opposite, or self-loop) of the edge from $n_k$ to $n_j$, $W^{dir(k,j)}$ is the direction-specific parameter, and $b^{L(k,j)}$ is the label-specific parameter.
Considering that syntactic information from all the neighboring nodes may make different contributions to semantic role labeling, the syntactic GCN introduces an additional edge-wise gate for each node pair ($n_k$, $n_j$) as $$g_{k,j} = \sigma(W_g^{dir(k,j)} x_k + b_g^{L(k,j)}).$$ The syntactic representation $v_k$ for a node $n_k$ can be then computed as: 
$$ v_k = \textit{ReLU}(\sum_{j \in N(k)} g_{k, j} \odot u_{k, j}).$$

\paragraph{SA-LSTM}\label{salstm}

The SA-LSTM \cite{qian-etal-2017-syntax} is an extension of the standard BiLSTM architecture, which aims to simultaneously encode the syntactic and contextual information for a given word. On the one hand, the SA-LSTM calculates the hidden state in timestep order as the standard LSTM,
\begin{gather*}
i_g = \sigma(W^{(i)} x_k + U^{(i)} h_{k-1} + b^{(i)}),\\
f_g = \sigma(W^{(f)} x_k + U^{(f)} h_{k-1} + b^{(f)}),\\
o_g = \sigma(W^{(o)} x_k + U^{(o)} h_{k-1} + b^{(o)}),\\
u = f(W^{(u)} x_k + U^{(u)} h_{k-1} + b^{(u)}),\\
c_k = i_g \odot u + f_g \odot c_{k-1}.
\end{gather*}

On the other hand, it further incorporates the syntactic information into the representation of each word by introducing an additional gate,
\begin{gather*}
s_g = \sigma(W^{(s)} x_k + U^{(s)} h_{k-1} + b^{(s)}),\\
h_k = o_g \odot f(c_k) + s_g \odot \tilde{h}_k.
\end{gather*} 
where $\tilde{h}_k = f( \sum_{t_j < t_k } \alpha_j \times h_j)$ is the weighted sum of all hidden state vectors $h_j$ that come from previous node (word) $n_j$, and the weight factor $\alpha_j$ is actually a trainable weight related to the dependency relation $L(k, \cdot)$ when there exists a directed edge from $n_j$ to $n_k$.

Note that $\tilde{h}_k$ is always the hidden state vector of the syntactic head of $n_k$ according to the definition of $\alpha_j$. Since a word will be assigned a single syntactic head, such a strict constraint prevents the SA-LSTM from incorporating complex syntactic structures. Inspired by the GCN, we relax the directed constraint of $\alpha_j$ whenever there is an edge between $n_j$ and $n_k$.

After the SA-LSTM transformation, the outputs of the SA-LSTM layer from both directions are concatenated and taken as the syntactic representation of each word $n_k$, i.e., $v_k = [\overrightarrow{h_k}, \overleftarrow{h_k}]$. Different from the syntactic GCN, SA-LSTM encodes both syntactic and contextual information in a single vector $v_k$.

\paragraph{Tree-LSTM}\label{treelstm}

The Tree-LSTM \cite{tai-etal-2015-improved} can be considered an extension of the standard LSTM and aims to model tree-structured topologies. At each timestep, it composes an input vector and the hidden states from arbitrarily many child units.  
Specifically, the main difference between the Tree-LSTM unit and the standard one is that the memory cell updating and the calculation of gating vectors are dependent on multiple child units. A Tree-LSTM unit can be connected to an arbitrary number of child units, and it assigns a single forget gate for each child unit. This provides Tree-LSTM the flexibility to incorporate or drop the information from each child unit.

Given a syntactic tree, the Tree-LSTM transformation is defined on node $n_k$ and its children set $C(k)$ and is formulated as follows \cite{tai-etal-2015-improved}:
\begin{gather}
\tilde{h}_k = \sum_{j \in C(k)} h_k,\label{eq1}\\
i_g = \sigma(W^{(i)}x_k + U^{(i)}\tilde{h}_k + b^{(i)}),\nonumber\\
f^{k, j}_g = \sigma(W^{(f)}x_k + U^{(f)}h_j + b^{(f)}),\label{eq2}\\ 
o_g = \sigma(W^{(o)}x_k + U^{(o)}\tilde{h}_k + b^{(o)}),\nonumber\\
u = \tanh(W^{(u)}x_k + U^{(u)}\tilde{h}_k + b^{(u)}),\nonumber\\
c_k = i_g \odot u + \sum_{j \in C(k)} f^{k, j}_g \odot c_j,\nonumber\\
h_k = o_g \odot \tanh(c_k).\nonumber
\end{gather}
where $j\in C(k)$, $h_j$ is the hidden state of the $j$-th child node, $c_k$ is the memory cell of the head node $k$, and $h_k$ is the hidden state of node $k$. Note that in Eq.(\ref{eq2}), a single forget gate $f^{k, j}_g$ is computed for each hidden state $h_j$.

Note that the primitive form of Tree-LSTM does not take the dependency relations into consideration. Given the importance of dependency relations in the SRL task, we further extend the Tree-LSTM by adding an additional gate $r_g$ and reformulate Eq. (\ref{eq1}),
\begin{gather*}
r_g^{k, j} = \sigma(W^{(r)} x_k + U^{(r)} h_j + b^{L(k, j)}),\\
\tilde{h}_k = \sum_{j \in C(k)} r_g^{k,j} \odot h_j.
\end{gather*}  
where $b^{L(k, j)}$ is a relation label-specific bias term. After the Tree-LSTM transformation, the hidden state of each node in the dependency tree is taken as its syntactic representation, i.e., $v_k = h_k$.

\subsection{Constituent Composition and Decomposition}

Due to the difference in structure between constituent and dependency syntax trees,  tree encoders (GCN, SA-LSTM, Tree-LSTM, etc.) cannot be used to encode the constituent tree directly. In order for the constituent syntax to be encoded into the SRL model as the dependency syntax tree was, inspired by \cite{marcheggiani2019graph}, we introduce two processes: constituent tree conversion and feature decomposition.

\begin{figure}
	\centering
	\includegraphics[width=1.0\textwidth]{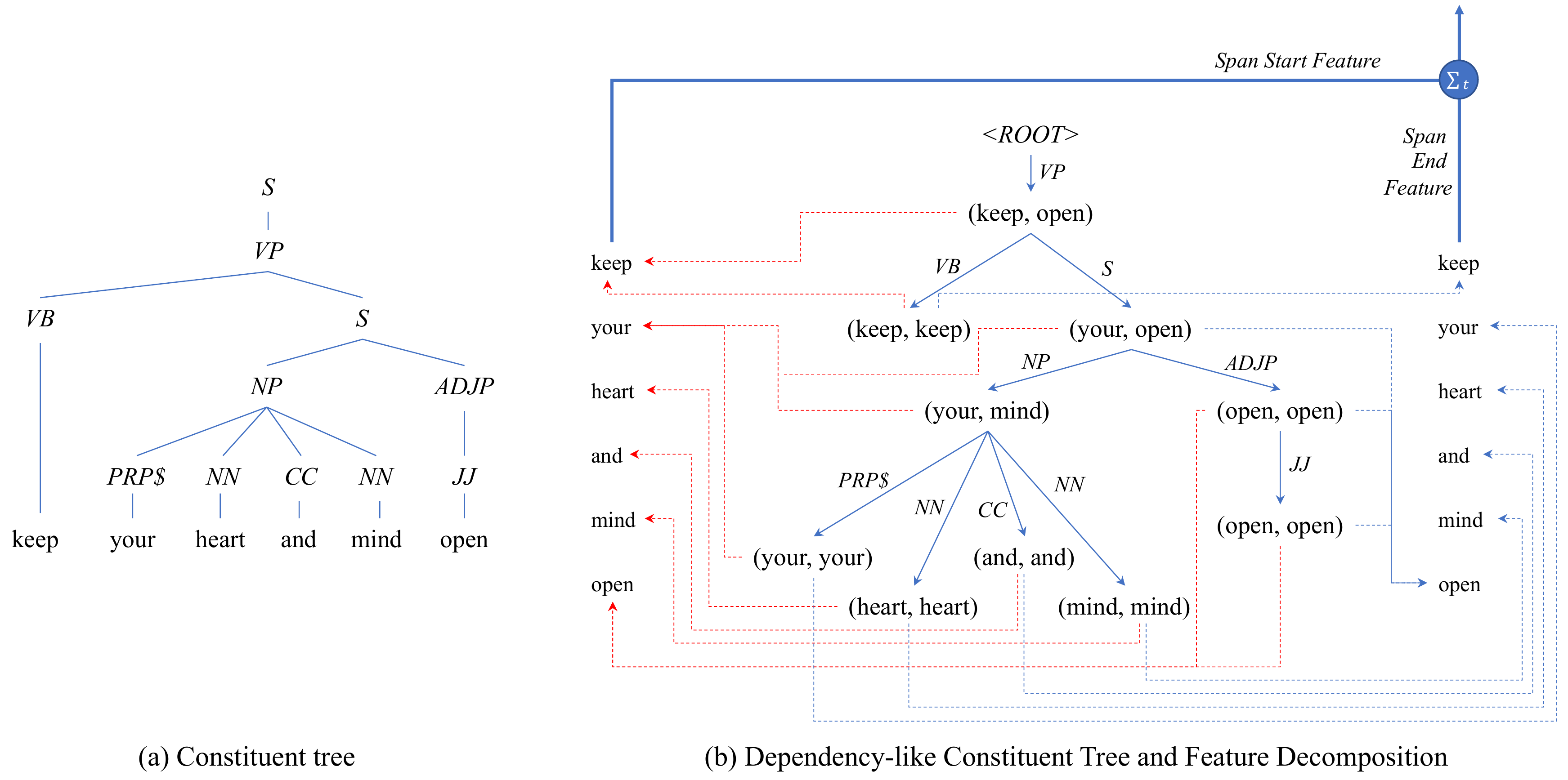}
	\caption{The structure of an original constituent tree and its dependency-like constituent tree after conversion. The dashed lines map the converted tree encoded features back to the original positions of words in the sentence, i.e., feature decomposition.} \label{fig:consituent_comp}
\end{figure}

A constituency tree is composed of terminal nodes and non-terminal nodes, as shown in Figure \ref{fig:consituent_comp}(a). Since the words in a constituent tree all are terminal nodes, if the constituent tree is directly encoded by the tree encoder, the syntax tree structural information cannot be encoded into the words fully. Therefore, we convert the constituent tree to a dependency-like tree, in which the original terminal nodes are removed and the remaining non-terminal nodes are replaced by units consisting of the start and end tokens (words) of the spans they represented. The constituent labels are modified to mimic dependency arcs as in dependency trees, as shown in Figure \ref{fig:consituent_comp}(b).

In a dependency tree, the nodes in the tree are the words in the sentence, so the syntax features output from the tree encoder can be mapped directly to the linear order of the sentences. In our converted constituent tree, the nodes in the tree correspond to the start and end words in the sentence, so we need an additional decomposition process to map this feature back to the word level. As shown in the dashed line in Figure \ref{fig:consituent_comp}(b), every node passes the feature to the first and the last words in their spans, and an extra indicator embedding $e^{spos}$ is appended to distinguish features as starts or ends. Then these features are input to tree encoders to obtain the final syntax features for each word.

\section{Experimental Analysis and Syntax Role Study} 

In this section, we investigate the proposed methods empirically in comparison to the latest SRL models. Moreover, we further explore the syntax role for neural SRL in various architectures.
The SRL models are evaluated on the popular CoNLL-2005, CoNLL-2009, and CoNLL-2012 shared tasks following the standard training, development and test splits. For the SRL task, because the predicate identification subtask is easier than other subtasks, some works only focus on the semantic role prediction with pre-identified predicates, which we name \textit{w/ pred}. There are also many studies that tend to examine settings closer to real-world scenarios, where the predicates are not given and the proposed systems are required to output both the predicates and their corresponding argument. We call this setting \textit{w/o pred}.

The hyperparameters in our model were selected based on the development set. In our experiments, all real vectors are randomly initialized, including 100-dimensional word, lemma, POS tag embeddings, and  16-dimensional predicate-specific indicator embeddings \cite{he-etal-2018-syntax}. The pre-trained word embeddings are 100-dimensional GloVe vectors \cite{pennington-etal-2014-glove} for English and 300-dimensional fastText vectors \cite{grave-etal-2018-learning} trained on Common Crawl and Wikipedia for other languages. The dimensions of ELMo and BERT word embeddings are of size 1024. Besides, we use a 3 layer BiLSTM with 400-dimensional hidden states and apply dropout with an 80\% keep probability between timesteps and layers. For the biaffine scorer, we employ two 300-dimensional affine transformations with the ReLU non-linear activation and also set the dropout probability to 0.2. During training, we use the categorical cross-entropy as the objective and use the Adam optimizer \cite{kingma2014adam} with an initial learning rate $2e^{-3}$. All models are trained for up to 50 epochs with batch size 64.

For the syntax input, we obtained the dependency syntax tree with Biaffine Parser \cite{dozat2017deep} in dependency SRL, while in span SRL, follow the practice of \cite{he-etal-2017-deep}, a leading constituency parser \cite{choe-charniak-2016-parsing} is used to provide constituent trees.

\subsection{Datasets}

\paragraph{CoNLL 2005 and 2012} The CoNLL-2005 shared task focused on verbal predicates only for English.
The CoNLL-2005 dataset takes sections 2-21 of Wall Street Journal (WSJ) data as the training set, and section 24 as the development set. The test set consists of section 23 of WSJ for in-domain evaluation together with 3 sections from the Brown corpus for out-of-domain evaluation. The larger CoNLL-2012 dataset is extracted from OntoNotes v5.0 corpus, which contains both verbal and nominal predicates.

\paragraph{CoNLL 2009}
The CoNLL-2009 shared task is focused on dependency-based SRL in multiple languages and merges two treebanks, PropBank and NomBank. NomBank is a complement to PropBank and uses a similar semantic convention for nominal predicate-argument structure annotation. The training, development, and test splits of the English data are identical to those of CoNLL-2005.

\subsection{Preprocessing}

\begin{figure}
	\centering
	\includegraphics[scale=0.66]{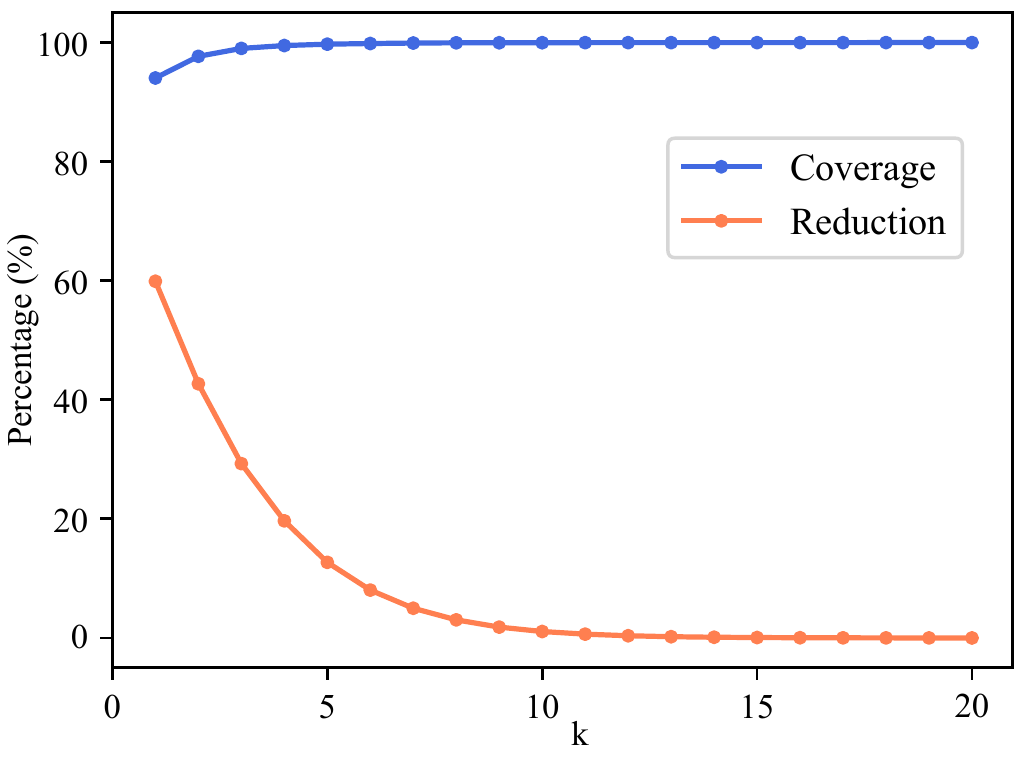}
	\caption{Changing curves of coverage and reduction with different $k$ values on the English training set. The coverage rate is the proportion of true arguments in the pruning output, while the reduction is the proportion of pruned argument candidates in total tokens.} \label{fig:k}
\end{figure}

\paragraph{Hard Pruning} During the pruning of argument candidates, we use the officially predicted syntactic parses provided by CoNLL-2009 shared-task organizers on both English and Chinese. Figure \ref{fig:k} shows changing curves of coverage and reduction following $k$ on the English train set. According to our statistics, the number of non-arguments is ten times more than that of arguments, meaning the data distribution is fairly unbalanced; however, a proper pruning strategy could alleviate this problem. Accordingly, the first-order pruning reduces more than 50\% candidates at the cost of missing 5.5\% true ones on average, and the second-order prunes about 40\% candidates with nearly 2.0\% loss. The coverage of third-order achieves 99\% , and it reduces the size of the corpus by approximately 1/3.

It is worth noting that when $k$ is larger than 19, full coverage is achieved on all argument candidates for the English training set, which allows our high order pruning algorithm to reduce to a syntax-agnostic setting. In this work, we use tenth-order pruning for best performance. 

\paragraph{Soft Pruning} For the syntactic rule used in soft argument pruning, to ensure more than 99\% coverage of true arguments in pruning output, we use the top-$120$ distance tuples on Japanese and top-$20$ on other languages for a better trade-off between computation and coverage.

\paragraph{Candidates Pruning}  For candidates pruning, we follow the settings of \citet{he-etal-2018-jointly}, modeling spans up to length $\mathcal{L}=30$ for span SRL and $\mathcal{L}=1$ for dependency SRL, using $\beta_p=0.4$ for pruning predicates and $\beta_a=0.8$ for pruning arguments.

\begin{table}
	\centering
	\caption{Dependency SRL Results with pre-identified predicates (\textit{w/ pred}) and without pre-identified predicates (\textit{w/o pred}) on the CoNLL-2009 English in-domain (WSJ) and out-of-domain (Brown) test sets. The "PLM" column indicates whether a and which pre-trained language model is used, the "SYN"  column indicates whether syntax information is employed, "+E" in the "PLM" column shows that the model leverages ELMo for pre-trained language model features.  \textbf{[Ens.]} is used to specify the ensemble system, \textbf{[Semi.]} indicates semi-supervised training is adopted, and \textbf{[Joint]}  means joint learning with other tasks.}\label{tab:resultE}
	\setlength{\tabcolsep}{3pt}
	\scalebox{0.72}{
		\begin{tabular}{lcccccccccccccc}
			\toprule
			\multirow{3}{*}{System} & \multirow{3}{*}{PLM} & \multirow{3}{*}{SYN} & \multicolumn{6}{c}{\textit{w/ pred}} & \multicolumn{6}{c}{\textit{w/o pred}} \\
			\cmidrule(lr){4-9} \cmidrule(lr){10-15} & & & \multicolumn{3}{c}{WSJ} & \multicolumn{3}{c}{Brown} & \multicolumn{3}{c}{WSJ} & \multicolumn{3}{c}{Brown}  \\
			\cmidrule(lr){4-6} \cmidrule(lr){7-9} \cmidrule(lr){10-12} \cmidrule(lr){13-15} &  & & P & R & F$_1$ & P & R & F$_1$ & P & R & F$_1$ & P & R & F$_1$ \\
			\midrule
			\cite{zhao-etal-2009-multilingual} & & Y & $-$ & $-$ & 85.4 & $-$ & $-$ & 73.3 & $-$  & $-$  & $-$  & $-$  & $-$  & $-$ \\
			\cite{zhao-etal-2009-multilingual-dependency} & & Y & $-$ & $-$ & 86.2 & $-$ & $-$ & 74.6 & $-$  & $-$  & $-$  & $-$  & $-$  & $-$ \\
			\cite{lei-etal-2015-high} & & Y & $-$ & $-$ & 86.6 & $-$ & $-$ & 75.6 & $-$  & $-$  & $-$  & $-$  & $-$  & $-$ \\
			\cite{fitzgerald-etal-2015-semantic} & & Y & $-$ & $-$ & 87.3 &$-$ & $-$ & 75.2 & $-$  & $-$  & $-$  & $-$  & $-$  & $-$ \\
			\textbf{[Ens.]} \cite{fitzgerald-etal-2015-semantic}  & & Y & $-$ & $-$ & 87.8 & $-$ & $-$ & 75.5 & $-$  & $-$  & $-$  & $-$  & $-$  & $-$ \\
			\cite{roth-lapata-2016-neural} & & Y & 90.0 & 85.5 & 87.7 & 78.6 & 73.8 & 76.1 & $-$  & $-$  & $-$  & $-$  & $-$  & $-$\\
			\textbf{[Ens.]} \cite{roth-lapata-2016-neural} & & Y & 90.3 & 85.7 & 87.9 & 79.7 & 73.6 & 76.5 & $-$  & $-$  & $-$  & $-$  & $-$  & $-$ \\
			\cite{swayamdipta-etal-2016-greedy} & & N & $-$ & $-$ & 85.0 & $-$ & $-$ & $-$ & $-$ & $-$ & 80.5 & $-$ & $-$ & $-$ \\
			\cite{marcheggiani-titov-2017-encoding} & & Y & 89.1 & 86.8 & 88.0 & 78.5 & 75.9 & 77.2 & $-$  & $-$  & $-$  & $-$  & $-$  & $-$ \\
			\textbf{[Ens.]} \cite{marcheggiani-titov-2017-encoding} & & Y & 90.5 & 87.7 & 89.1 & 80.8 & 77.1 & 78.9 & $-$  & $-$  & $-$  & $-$  & $-$  & $-$ \\
			\cite{marcheggiani-etal-2017-simple} & & N & 88.7 & 86.8 & 87.7 & 79.4 & 76.2 & 77.7 & $-$  & $-$  & $-$  & $-$  & $-$  & $-$ \\
			\cite{mulcaire-etal-2018-polyglot} & & N & $-$ & $-$ & 87.2 & $-$  & $-$  & $-$  & $-$  & $-$  & $-$ & $-$ & $-$ & $-$\\ \hdashline
			\multirow{2}{*}{\cite{kasai-etal-2019-syntax}} & & Y & 89.0 & 88.2 & 88.6 & 78.0 & 77.2 & 77.6 & $-$  & $-$  & $-$  & $-$  & $-$  & $-$ \\
			& +E & Y & 90.3 & 90.0 & 90.2 & 81.0 & 80.5 & 80.8 & $-$  & $-$  & $-$  & $-$  & $-$  & $-$ \\ \hdashline
			\cite{cai-lapata-2019-semi} & & N & 91.1 & 90.4 & 90.7 & 82.1 & 81.3 & 81.6 & $-$  & $-$  & $-$  & $-$  & $-$  & $-$ \\
			\textbf{[Semi.]}  \cite{cai-lapata-2019-semi} & & N & 91.7 & 90.8 & \bf 91.2 & 83.2 & 81.9 & 82.5 & $-$  & $-$  & $-$  & $-$  & $-$  & $-$ \\
			\cite{zhang-etal-2019-syntax-enhanced} & & Y & 89.6 & 86.0 &  87.7 & $-$  & $-$  & $-$  & $-$  & $-$  & $-$ & $-$  & $-$  & $-$ \\
			\cite{lyu-etal-2019-semantic} & +E & N & $-$ & $-$ & 91.0 & $-$ & $-$ & 82.2 & $-$  & $-$  & $-$  & $-$  & $-$  & $-$ \\
			\cite{chen-etal-2019-capturing} & +E & N  & 90.7 & 91.4 & 91.1 & 82.7 & 82.8 & 82.7  & $-$  & $-$  & $-$  & $-$  & $-$  & $-$  \\ \hdashline
			\multirow{2}{*}{\textbf{[Joint]} \cite{zhou2019parsing}} & & N & 88.7 & 89.8 & 89.3 & 82.5 & 83.2 & 82.8 & 84.2 & 87.6 & 85.9 & 76.5 & 78.5 & 77.5 \\
			 & +E & N & 89.7 & 90.9 & 90.3 & 83.9 & 85.0 & \bf 84.5 & 85.2 & 88.2 & \bf 86.7 & 78.6 & 80.8 & \bf  79.7 \\
			\midrule
			\textbf{Sequence-based} (\citeyear{he-etal-2018-syntax, li-etal-2018-unified}) &  +E & N & 89.5 & 87.9 & 88.7 & 81.7 & 76.1 & 78.8 & 83.5 & 82.4 & 82.9 & 71.5 & 70.9 & 71.2 \\
			\quad\textbf{+K-order Hard Pruning} (\citeyear{he-etal-2018-syntax}) & +E & Y& 89.7 & 89.3 & 89.5 & 81.9 & 76.9 & 79.3 & 83.9 & 82.7 & 83.3 & 71.5 & 71.3 & 71.4 \\
			\quad \textbf{+SynRule Soft Pruning}  & +E & Y & 89.9  & 89.1  & 89.5 & 78.8 & 81.2 & \bf 80.0 & 82.9 & 84.3 & 83.6 & 70.9 & 72.1 & 71.5 \\
			\quad \textbf{+GCN Syntax Encoder} (\citeyear{li-etal-2018-unified}) & +E & Y & 90.3  & 89.3  & \bf 89.8 & 80.6 & 79.0 & 79.8 & 85.3 & 82.5 & 83.9 & 71.9 &	71.5 &	\bf 71.7\\
			\quad \textbf{+SA-LSTM Syntax Encoder} (\citeyear{li-etal-2018-unified}) & +E & Y & 90.8  & 88.6  & 89.7 & 81.0 & 78.2 & 79.6 & 85.3 &	82.6	& \bf 84.0 & 71.8 & 71.6 &	\bf 71.7 \\
			\quad \textbf{+Tree-LSTM Syntax Encoder} (\citeyear{li-etal-2018-unified}) & +E & Y & 90.0  & 88.8 & 89.4 & 80.4 & 78.7 & 79.5 & 83.1 & 83.7 &	83.4 & 70.9 & 72.1 & 71.5\\
			\midrule
			\textbf{Tree-based}  (\citeyear{cai-etal-2018-full}) & +E & N & 89.2 &  90.4 & 89.8 & 80.0 & 78.6 &  79.3  & 84.8 & 85.4 & 85.1 & 72.4 & 74.0 & 73.2 \\
			\quad\textbf{+K-order Hard Pruning}  & +E & Y &  90.3 & 89.5 & 89.9 & 80.0 & 79.0 & 79.5 & 83.9	& 86.5 &	85.2 & 73.6 & 72.8 & 73.2\\
			\quad \textbf{+SynRule Soft Pruning} (\citeyear{he-etal-2019-syntax}) & +E & Y & 90.0  & 90.7  & 90.3 & 79.6 & 80.4 & 80.0 & 84.9 & 85.9 & 85.4 & 72.7 & 74.3 & 73.5 \\
			\quad \textbf{+GCN Syntax Encoder} & +E & Y &  90.9 & 90.1 & \bf 90.5 & 81.4 & 78.8 & 80.1 & 86.1 & 84.9 & \bf 85.5 & 73.5 & 73.7 & \bf 73.6\\
			\quad \textbf{+SA-LSTM Syntax Encoder} & +E & Y & 91.1 & 89.9 & \bf 90.5 & 80.9 & 79.5 & \bf 80.2 & 85.3 & 85.0	& 85.2 & 72.9 & 73.5 & 73.2\\
			\quad \textbf{+Tree-LSTM Syntax Encoder} & +E & Y &  89.8 & 90.6 & 90.2 & 80.0 & 79.8 & 79.9 & 85.3 & 85.3  & 85.3 & 73.9 & 73.1 & 73.5\\
			\midrule
			\textbf{Graph-based} (\citeyear{li2019dependency}) & +E & N & 90.0 & 90.0 & 90.0 & 81.7  & 81.4 & 81.5  & 85.6 & 85.0 & 85.3 & 73.0 & 74.0 & 73.5 \\
			\quad\textbf{+K-order Hard Pruning}  & +E & Y &  90.3 & 89.7 & 90.0 & 80.7 & 81.9 & 81.3 & 84.6 & 85.8 & 85.2 & 73.7 & 73.3 & 73.5 \\
			\quad \textbf{+SynRule Soft Pruning} & +E & Y &  89.8 & 90.6 & 90.2 & 80.8 & 82.4 & 81.6 & 85.0 & 86.0 & 85.5 & 72.8 & 74.4 & 73.6 \\
			\quad \textbf{+GCN Syntax Encoder} & +E & Y &  90.5 & 91.7 & \bf 91.1 & 83.3 & 80.9 & \bf 82.1 & 86.2 & 86.0 & \bf  86.1 & 73.8 & 74.6 & \bf 74.2 \\
			\quad \textbf{+SA-LSTM Syntax Encoder} & +E & Y & 91.0 & 90.4 & 90.7 & 82.4 & 81.6 & 82.0 & 86.3 & 85.5 & 85.9 & 75.4 & 72.8 & 74.1 \\
			\quad \textbf{+Tree-LSTM Syntax Encoder} & +E & Y &  90.7 & 90.3 & 90.5 & 80.2 & 83.4 & 81.8 & 86.9 & 84.3 & 85.6 & 74.1 & 73.7 & 73.9 \\
			\bottomrule
		\end{tabular}
	}
	
\end{table}

\subsection{Dependency SRL Results}

Undoubtedly, dependency SRL offers a number of advantages from a practical perspective, and the efficient dependency parsing algorithms enable SRL models to achieve state-of-the-art results.
Therefore, we begin our exploration of syntax roles for neural SRL with it. In Table \ref{tab:resultE}, we outlined the performance of the current leading dependency SRL models and compared the performance of our three baselines and syntax-enhanced models with different integration approaches on the CoNLL-2009 English in-domain (WSJ) and out-of-domain (Brown) test sets.

In the sequence-based approaches, we employed another sequence labeling model to tackle the predicate identification and disambiguation subtasks required for the different settings (\textit{w/ pred} and \textit{w/o pred}). The predicate disambiguation model achieves accuracies of 95.01\% and 95.58\% on the development and test (WSJ) sets for the \textit{w/ pred} setting, respectively, giving a slightly better accuracy than \cite{roth-lapata-2016-neural}, which had 94.77\% and 95.47\% accuracy on development and test sets, respectively. As for the \textit{w/o pred} setting, the $F_1$ score of our predicate labeling model is 90.11\% and 90.53\% on development and test (WSJ) sets, respectively. With the help of the ELMo pre-trained language model, our sequence-based baseline model has achieved competitive results when compared to the other leading SRL models. Compared to the closest work \cite{marcheggiani-etal-2017-simple}, ELMo brought a 1.0\% improvement for our baseline model, which verifies it is a strong baseline. In this case, the syntax enhancement gave us a performance improvement of 0.8\%-1.1\% (\textit{w/ pred}, in-domain test set), demonstrating that both hard/soft pruning and syntax encoders effectively exploit syntax for sequence-based neural SRL models.

In the tree-based approaches, the predicate disambiguation subtask is unifiedly tackled with argument labeling by making predictions on the $\text{ROOT}$ node of the factorized tree. The disambiguation precision is 95.0\% in the \textit{w/ pred} setting, while in the \textit{w/o pred} setting, we first attach all the words in the sentence to the $\text{ROOT}$ node and label the word that is not a predicate with the $null$ role label. It should be noted that in the \textit{w/o pred} setting, we just attach the predicates to the $\text{ROOT}$ node, since we do not need to distinguish the predicate from other words. The training scheme remains the same as in the \textit{w/o pred} setting, while in the inference phase, an additional procedure is performed to find out all the predicates of a given sentence. The $F_1$ score on predicates identification and labeling of this process is 89.43\%. Based on the tree-based baseline, the hard pruning syntax enhancement fails to improve on the baseline despite the hard pruning method's ability to alleviate the imbalanced label distribution caused by the $null$ role labels. We suspect the possible reason is the employment of a biaffine attention structure, which already alleviates imbalanced label distribution issue. This is problematic because both the biaffine attention and hard pruning work to balance the label distribution, and after the biaffine attention balances it to some degree, hard pruning is much more likely to incorrectly prune true labels, which potentially even leads to a decrease in performance. Compared with hard pruning, soft pruning can greatly reduce the incorrect pruning of true arguments, which serve as clues in the model. Because of this, the soft pruning algorithm applied to the tree-based model can obtain performance improvement similar to that of the tree-based baseline. In addition, the performance improvements of syntax encoders in the tree-based model are similar to those of the sequence-based model.

In the graph-based approaches, because the introduction of candidate pruning, argument pruning is directly controlled by the neural network scorer, both syntax-based hard and soft pruning methods lose the effects they provided in the sequence-based and tree-based models. The syntax encoder, however, can provide quite stable performance improvement as it does in sequence-based and tree-based models.

\begin{table*}[t!]
	\centering
	\caption{Dependency SRL results on the CoNLL-2009 multilingual in-domain test sets with pre-identified predicates (\textit{w/ pred}) setting. The first row is the best result of the CoNLL-2009 shared task \cite{hajic-etal-2009-conll}. The "PLM" column indicates whether and which pre-trained language models is used, the "SYN"  column indicates whether syntax information is employed, "+E" in the "PLM" column indicates the model leverages pre-trained ELMo features for all languages, "+E$^\dagger$" in the "PLM" column represents ELMo is only used for English.}\label{tab:all-results}
	\scalebox{0.75}{
		\begin{tabular}{lcccccccccc}
			\toprule
			System & PLM & SYN & CA  & CS & DE & EN & ES &  JA & ZH & Avg. \\
			\midrule
			CoNLL-2009 best  & & Y & 80.3  & 86.5 & 79.7 & 86.2  & 80.5 & 78.3  & 78.6 & 81.4 \\
			\cite{zhao-etal-2009-multilingual-dependency} & & Y & 80.3 & 85.2 & 76.0 & 85.4 & 80.5 & 78.2 & 77.7  & 80.5 \\
			\cite{roth-lapata-2016-neural} & & Y & $-$  & $-$ & 80.1 & 87.7 & 80.2 & $-$ & 79.4 & $-$ \\
			\cite{marcheggiani-titov-2017-encoding} & & Y & $-$  & $-$  & $-$  & 88.0 & $-$  & $-$  & 82.5 & $-$  \\
			\cite{marcheggiani-etal-2017-simple} & & N & $-$  & 86.0 & $-$ & 87.7  & 80.3 & $-$ & 81.2 & $-$ \\
			\cite{mulcaire-etal-2018-polyglot} & & N & 79.5 & 85.1 & 70.0 & 87.2 & 77.3 & 76.0 & 81.9 & 79.6 \\
			\cite{kasai-etal-2019-syntax} & +E & Y & $-$  & $-$ & $-$& 90.2 & \bf 83.0 & $-$ & $-$ & \\
			\cite{cai-lapata-2019-semi} & & N & $-$ & $-$ & 83.3 & 90.7 & 82.1 & $-$ & 84.6 & $-$ \\
			\textbf{[Semi.]} \cite{cai-lapata-2019-semi} & & N & $-$ & $-$ & \bf 83.8 & \bf 91.2 & 82.9 & $-$ & \bf 85.0 & $-$ \\
			\cite{zhang-etal-2019-syntax-enhanced} & & Y & $-$ & $-$ & $-$ & 87.7 & $-$ & $-$ & 84.2 & $-$ \\
			\cite{lyu-etal-2019-semantic} & $\ $+E$^\dagger$ & N & 80.9  & 87.6 & 75.9  & 91.0 & 80.5 & \bf 82.5  & 83.3 & \bf 83.1 \\
			\cite{chen-etal-2019-capturing} & $\ $+E$^\dagger$ & N & \bf 81.7 & \bf 88.1 & 76.4 & 91.1 & 81.3 & 81.3 & 81.7 & \bf 83.1 \\
			\midrule
			\textbf{Sequence-based} (\citeyear{he-etal-2018-syntax, li-etal-2018-unified}) & +E & N & 84.0 &  87.8 & 76.8 & 88.7 & 82.9 & 82.8 & 83.1 & 83.7 \\
			\quad\textbf{+K-order Hard Pruning} (\citeyear{he-etal-2018-syntax})  & +E & Y & 84.5 & 88.3 & 77.3 & 89.5 & 83.3 & 82.9 & 82.8 & 84.1  \\
			\quad \textbf{+SynRule Soft Pruning} & +E & Y  & 84.4 & 88.2 & \bf 77.5 &  89.5 & 83.2 & 83.0 & 83.3 & 84.2  \\
			\quad \textbf{+GCN Syntax Encoder}  (\citeyear{li-etal-2018-unified}) & +E & Y & \bf 84.6 & \bf 88.5 & 77.2 & \bf 89.8 & \bf 83.6 & \bf 83.2 & \bf 83.8 & 84.4 \\
			\quad \textbf{+SA-LSTM Syntax Encoder}  (\citeyear{li-etal-2018-unified}) & +E & Y & 84.3  & \bf 88.5 & 77.0 & 89.7 & 83.5 & 83.1 & 83.5 & 84.2 \\
			\quad \textbf{+Tree-LSTM Syntax Encoder}  (\citeyear{li-etal-2018-unified}) & +E & Y & 84.1 & 88.3 & 76.9 & 89.4 & 83.2 & 82.9 & 83.4 & 84.0 \\
			\midrule
			\textbf{Tree-based} (\citeyear{cai-etal-2018-full}) & +E & N & 84.1 & 88.4 & 78.4 & 89.9 & 83.5 & 83.0 & 84.0 & 84.5 \\
			\quad\textbf{+K-order Hard Pruning}  & +E & Y & 84.2 & 88.5 & 78.4 & 89.9  & 83.4 & 82.8 & 84.2 & 84.5 \\
			\quad\textbf{+SynRule Soft Pruning} (\citeyear{he-etal-2019-syntax}) & +E & Y & 84.4  & 88.8 & 78.5 & 90.0  & 83.7 & 83.1  & 84.6 & 84.7 \\
			\quad \textbf{+GCN Syntax Encoder} & +E & Y & \bf 84.8 & \bf 89.3 & 78.1 & \bf 90.2 & \bf 84.0 & \bf 83.3 & \bf 85.0 & \bf 85.0 \\
			\quad \textbf{+SA-LSTM Syntax Encoder} & +E & Y  & 84.6  & 89.0 & \bf 78.8 & 90.0 & 83.7 & 83.1 & 84.8 & 84.9\\
			\quad \textbf{+Tree-LSTM Syntax Encoder} & +E & Y  & 84.4 & 88.9 & 78.6 & 89.9 & 83.6 & 83.0 & 84.5 & 84.7 \\
			\midrule
			\textbf{Graph-based} (\citeyear{li2019dependency}) & +E & N & 85.0 & 90.2 & 76.0 & 90.0 & 83.8 & 82.7 & 85.7 & 84.8 \\
			\quad\textbf{+K-order Hard Pruning}  & +E & Y  & 84.9 & 90.2 & 75.7 & 89.8 & 83.5  & 82.8 & 85.8 & 84.7 \\
			\quad \textbf{+SynRule Soft Pruning} & +E & Y & 85.2 & 90.3 &  76.2 & 90.1 & 84.0  & 82.9 & 85.8 & 84.9 \\
			\quad \textbf{+GCN Syntax Encoder} & +E & Y & \bf 85.5 & \bf 90.5 & \bf 76.6 & \bf 90.4 & \bf 84.3 & \bf 83.2 & \bf 86.1 & \bf 85.2 \\
			\quad \textbf{+SA-LSTM Syntax Encoder} & +E & Y & 85.2 & \bf 90.5 & 76.4  & 90.3 & 84.1 & \bf 83.2 & 86.0  & 85.1 \\
			\quad \textbf{+Tree-LSTM Syntax Encoder} & +E & Y &  85.0 & 90.3 & 76.2 & 90.3 & 84.0 & 83.0 & 85.8 & 84.9 \\
			\bottomrule
		\end{tabular}
	}
	
\end{table*}

Though most SRL literature is dedicated to impressive performance gains on the English benchmark, exploring syntax's enhancing effects on diverse languages is also important for examining the role of syntax in SRL. Table \ref{tab:all-results} presents all in-domain test results on seven languages of CoNLL-2009 datasets. Compared with previous methods, our baseline yields strong performance on all datasets. Nevertheless, applying the syntax information to the strong syntax-agnostic baseline can still boost the model performance in general, which demonstrates the effectiveness of syntax information. On the other hand, the similar performance impact of hard/soft argument pruning on the baseline models indicates that syntax is generally beneficial to multiple languages and can enhance multilingual SRL performance with effective syntactic integration.

Based on the above results and analysis, we conclude that the syntax-based pruning algorithms proposed before the era of neural network can still play a role under certain conditions in the neural network era; however, when there are neural structures whose motivation is consistent with the original intention of these syntax-based pruning algorithms emerges, the effects of these algorithms are quite limited, and they can even have negative effects. Despite this limitation, the neural syntax encoder is beneficial in that it delegates the decisions of how to include syntax and what kind of syntax to use to neural networks, which reduces the number of manually defined features. This is an important result of the transition from the pre-neural network era to the neural network era. Handcrafted features may be advantageous when compared to poorly designed neural networks, but well-designed neural networks can easily outperform models relying on handcrafted features. In addition, neural models can also benefit from handcrafted features, so the two do not necessarily have to be directly compared, even though neural networks reduce the need for handcrafted features.

\subsection{Span SRL Results}

Apart from the dependency SRL experiments, we also conducted experiments to compare different syntax utilization on span SRL models. Table \ref{tab:gold-for-span} shows results on the CoNLL-2005 in-domain (WSJ) and out-of-domain (Brown) test sets, as well as the CoNLL-2012 test set (OntoNotes). The first block of the table presents results from previous works. These results demonstrate that with the development of neural networks, in particular the emergence of pre-trained language models, SRL achieved a large performance increase of more than 8.0\%, and syntax further enhanced the effect of these strong baselines, enabling syntax+pre-trained language models to achieve the state-of-the-art results \cite{wang2019best}. This indicates that the effect of SRL can still be improved as long as the syntax is used properly under current circumstances.

Besides, comparing the results of \cite{strubell-etal-2018-linguistically} and \cite{he-etal-2018-jointly}, it can be found that the feature extraction ability of self-attention is stronger than that of RNN, and the self-attention baseline obviously outperforms RNN-based one, but when syntactic information or a pre-trained language model is used to enhance performance, the performance margin becomes smaller. Therefore, we can speculate that self-attention implicitly and partially functions as the syntax information, as do pre-trained language models. 

\begin{table*}
	\centering
	\caption{Span SRL results with pre-identified predicates on the CoNLL-2005 and CoNLL-2012 test sets. The "PLM" column indicates whether a and which pre-trained language model is used, the "SYN"  column indicates whether syntax information is employed, "+E" in the "PLM" column shows that the model leverages the ELMo for pre-trained language model features.  \textbf{[Ens.]} is used to specify the ensemble system and \textbf{[Joint]}  means joint learning with other tasks.}\label{tab:gold-for-span}
	\setlength{\tabcolsep}{3pt}
	\scalebox{0.75}{
		\begin{tabular}{lccccccccccc}  
			\toprule  
			\multirow{2}{*}{System} & \multirow{2}{*}{PLM} & \multirow{2}{*}{SYN}  &\multicolumn{3}{c}{CoNLL05 WSJ}&\multicolumn{3}{c}{CoNLL05 Brown}&\multicolumn{3}{c}{CoNLL12}\\
			\cmidrule(lr){4-7} \cmidrule(lr){7-9} \cmidrule(lr){10-12} && &P&R&F$_1$&P&R&F$_1$&P&R&F$_1$\\  
			\midrule
			\textbf{[Ens.]} \cite{punyakanok-etal-2008-importance} & & Y &82.3&76.8&79.4&73.4&62.9&67.8&$-$&$-$&$-$ \\
			\cite{toutanova-etal-2008-global} & & Y & $-$ & $-$ & 79.7 & $-$ & $-$ & 67.8 & $-$ & $-$ & $-$\\
			\textbf{[Ens.]} \cite{toutanova-etal-2008-global} & & Y &81.9&78.8&80.3&$-$&$-$&68.8&$-$&$-$&$-$ \\
			\cite{pradhan-etal-2013-towards}$^*$ & & Y &$-$&$-$&$-$&$-$&$-$&$-$&78.5&76.6&77.5\\
			\cite{tackstrom-etal-2015-efficient} & & Y &82.3 & 77.6 & 79.9& 74.3 & 68.6 & 71.3&80.6 & 78.2 & 79.4 \\
			\cite{zhou-xu-2015-end} & & N & 82.9&82.8&82.8&70.7&68.2&69.4&$-$&$-$&81.3\\
			\cite{fitzgerald-etal-2015-semantic} & & Y & 81.8 & 77.3 & 79.4 & 73.8 & 68.8 & 71.2 & 80.9 & 78.4 & 79.6 \\
			\textbf{[Ens.]} \cite{fitzgerald-etal-2015-semantic} & & Y &82.5&78.2&80.3&74.5&70.0&72.2&81.2&79.0&80.1\\
			\cite{he-etal-2017-deep} & & Y &83.1&83.0&83.1&72.9&71.4&72.1&81.7&81.6&81.7\\
			\textbf{[Ens.]} \cite{he-etal-2017-deep} & & Y & 85.0 & 84.3 & 84.6 & 74.9 & 72.4 & 73.6 & 83.5 & 83.3 & 83.4 \\
			\cite{yang-mitchell-2017-joint} & & N &$-$&$-$&81.9&$-$&$-$&72.0&$-$&$-$&$-$ \\
			\cite{selfatt2018} & & N &84.5&85.2&84.8&73.5&74.6&74.1&81.9&83.6&82.7\\
			\textbf{[Ens.]} \cite{selfatt2018} & & N & 85.9 & 86.3 & 86.1 & 74.6 & 75.0 & 74.8 & 83.3 & 84.5 & 83.9\\ \hdashline
			\multirow{2}{*}{\cite{peters-etal-2018-deep}} &  & N &$-$&$-$&$-$&$-$&$-$&$-$&$-$&$-$&81.4\\
			& +E & N &$-$&$-$&$-$&$-$&$-$&$-$&$-$&$-$& 84.6\\ \hdashline
			\multirow{2}{*}{\cite{he-etal-2018-jointly}} &  & N &$-$&$-$&83.9&$-$&$-$&73.7&$-$&$-$&82.1\\
			& +E & N &$-$&$-$&87.4&$-$&$-$& \bf 80.4 &$-$&$-$&85.5 \\ \hdashline
			\multirow{2}{*}{\cite{strubell-etal-2018-linguistically}} & & N & 84.7 & 84.2 & 84.5 & 73.9 & 72.4 & 73.1 &$-$&$-$ &$-$ \\
			& & Y & 84.6 & 84.6 &  84.6 & 74.8 & 74.3 & 74.6 &$-$&$-$&$-$\\ \hdashline
			\multirow{2}{*}{\cite{ouchi-etal-2018-span}} & & N & 84.7 & 82.3 & 83.5 & 76.0 & 70.4 & 73.1 & 84.4 & 81.7 & 83.0 \\
			& +E & N & 88.2 & 87.0 & 87.6 & 79.9 & 77.5 & 78.7 & 87.1 & 85.3 & 86.2 \\ \hdashline
			\multirow{2}{*}{\cite{wang2019best}} & +E & N &$-$ &$-$ & 87.7 &$-$ &$-$ & 78.1 &$-$ &$-$ & 85.8 \\
			& +E & Y &$-$ &$-$ & \bf 88.2 &$-$ &$-$ & 79.3 &$-$ &$-$ & \bf 86.4 \\ \hdashline
			\cite{marcheggiani2019graph} & & Y & 85.8 & 85.1 & 85.4 & 76.2 & 74.7 & 75.5 & 84.5 & 84.3 & 84.4 \\ \hdashline
			\multirow{2}{*}{\textbf{[Joint]} \cite{zhou2019parsing}} & & N & 85.9 & 85.8 & 85.8 & 76.9 & 74.6 & 75.7 & $-$ & $-$ & $-$\\
			& +E & N & 87.8 & 88.3 & 88.0 & 79.6 & 78.6 & 79.1 & $-$ & $-$ & $-$\\
			\midrule
			\textbf{Sequence-based} & +E & N & 87.4 & 	85.6 & 86.5 & 80.0 	& 78.1  & 79.0 &  84.2 & 85.6 &  84.9 \\
			\quad \textbf{+GCN Syntax Encoder} & +E & Y & 87.2  & 86.8 & \bf 87.0 & 78.6 &	80.2 &  \bf 79.4 & 85.3 & 85.7 & \bf 85.5 \\
			\quad \textbf{+SA-LSTM Syntax Encoder} & +E & Y & 87.1 & 86.5  & 86.8 & 79.3  & 78.9 & 79.1 & 85.9  & 84.3 & 85.1 \\
			\quad \textbf{+Tree-LSTM Syntax Encoder} & +E & Y & 87.1  & 85.9  & 86.5 & 78.8 &	79.2 & 79.0 & 85.2 	&	84.2  & 84.7 \\
			\midrule
			\textbf{Tree-based} & +E & N & 88.8 &	86.0  & 87.4 & 79.9 & 79.5  &79.7  & 86.6 &	84.8 &  85.7 \\
			\quad \textbf{+GCN Syntax Encoder} & +E & Y & 	87.7 & 88.3  & \bf 88.0 & 81.1 	& 79.9 & \bf 80.5 & 86.9 & 85.5 & \bf 86.2 \\
			\quad \textbf{+SA-LSTM Syntax Encoder} & +E & Y &  87.5 &	87.6  & 87.6 & 80.4 & 79.8  & 80.1& 86.3 & 85.3  &  85.8 \\
			\quad \textbf{+Tree-LSTM Syntax Encoder} & +E & Y & 87.0 & 87.6  & 87.3 & 81.0 & 79.1  & 80.0 & 86.0 	& 85.6  & 85.8 \\
			\midrule
			\textbf{Graph-based} (\citeyear{li2019dependency}) & +E & N & 87.9 & 87.5 & 87.7 & 80.6 & 80.4 & 80.5 & 85.7 & 86.3 & 86.0 \\
			\quad\textbf{+Constituent Soft Pruning}  & +E & Y & 88.4  & 87.4  &  87.9 & 80.9 & 80.3  & 80.6 & 85.5 	& 86.9  & 86.2  \\
			\quad \textbf{+GCN Syntax Encoder} & +E & Y & 89.0 & 88.2 & \bf 88.6 & 80.8 & 81.2 & 81.0 & 87.2 & 86.2 & \bf 86.7 \\
			\quad \textbf{+SA-LSTM Syntax Encoder} & +E & Y &  88.6 & 87.8 & 88.2 & 81.0  & 81.2 & \bf 81.1 & 87.0  & 85.8 & 86.4 \\
			\quad \textbf{+Tree-LSTM Syntax Encoder} & +E & Y & 86.9 &	89.1  & 88.0 & 81.5 & 80.3& 80.9 & 86.6 & 86.0 & 86.3 \\
			\bottomrule  
		\end{tabular}
		}
\end{table*}
	
By comparing our full model to state-of-the-art SRL systems, we show that our model genuinely benefits from incorporating syntactic information and other modeling factorization. Although our use of constituent syntax requires the composition and decomposition processes, which contrasts the simple and intuitive dependency syntax, we achieved consistent improvements compared to dependency SRL on all three baselines: sequence-based, tree-based and graph-based. This shows that these syntax encoders are general for syntax choice and can encode syntax effectively.
	
Constituent syntax information is usually used in span SRL, while the dependency tree is adopted for the argument pruning algorithm. Additionally, in the sequence-based and tree-based factorizations of span SRL, argument spans are linearized with multiple B-, I-, and O tags, which alleviates some label imbalance problems and hence lessens the need for argument pruning. Constituent syntax mainly provides the boundary information of span for the model to guide the model to predict the correct argument span when it is used for pruning. In graph-based models, because the argument span exists alone, the boundary set obtained by the constituent tree can be used to prune candidate arguments. As shown in the results, constituent-based soft pruning can still improve performance on the graph-based baseline, but the improvement is smaller than that of syntax encoders, indicating that the syntax encoder can extract more information than just span boundaries.

\begin{table*}
	\centering
	\caption{Span SRL results without pre-identified predicates on the CoNLL-2005 and CoNLL-2012 data sets. The "PLM" column indicates whether and which pre-trained language model is used, the "SYN"  column indicates whether syntax information is employed, "+E" in the "PLM" column shows that the model leverages the ELMo for pre-trained language model features.  \textbf{[Ens.]} is used to specify the ensemble system and \textbf{[Joint]}  means joint learning with other tasks.}\label{tab:end-for-span}
	\scalebox{0.75}{
		\begin{tabular}{lccccccccccc}  
			\toprule  
			\multirow{2}{*}{System}  & \multirow{2}{*}{PLM} & \multirow{2}{*}{SYN} &\multicolumn{3}{c}{CoNLL05 WSJ}&\multicolumn{3}{c}{CoNLL05 Brown}&\multicolumn{3}{c}{CoNLL12}\\
			\cmidrule(lr){4-6} \cmidrule(lr){7-9} \cmidrule(lr){10-12}
			&&&P&R&F$_1$&P&R&F$_1$&P&R&F$_1$\\
			\midrule 
			\cite{he-etal-2017-deep} & & N &80.2&82.3&81.2&67.6&69.6&68.5&78.6&75.1&76.8\\
			\textbf{[Ens.]} \cite{he-etal-2017-deep}  & & N & 82.0&83.4&82.7&69.7&70.5&70.1&80.2&76.6&78.4\\
			\cite{he-etal-2018-jointly} & +E & N &84.8&87.2&86.0&73.9& 78.4 &76.1&81.9& 84.0 &82.9\\  \hdashline
			\multirow{2}{*}{\cite{strubell-etal-2018-linguistically}} &  & Y & 84.0 & 83.2 & 83.6 & 73.3 & 70.6 & 71.9 & 81.9 & 79.6 & 80.7\\
			& +E & Y & 86.7 & 86.4 & \bf 86.6 & 79.0 & 77.2 & \bf 78.1 & 84.0 & 82.3 & \bf 83.1\\ \hdashline
			\multirow{2}{*}{\textbf{[Joint]} \cite{zhou2019parsing}} & & N & 83.7 & 85.5 & 84.6 & 72.0 & 73.1 & 72.6 & $-$ & $-$  & $-$  \\
			& +E & N & 85.3 & 87.7 & 86.5 & 76.1 & 78.3 & 77.2 & $-$ & $-$  & $-$  \\
			\midrule
			\textbf{Sequence-based} & +E  & N & 84.4 & 83.6 & 84.0 & 76.5 & 73.9 & 75.2 & 81.7 & 82.9 &  82.3 \\
			\quad \textbf{+GCN Syntax Encoder} & +E  & Y & 85.5 &	84.3 & \bf 84.9 & 78.8 	& 73.4 & \bf 76.0 & 83.1 & 82.5  & \bf 82.8 \\
			\quad \textbf{+SA-LSTM Syntax Encoder} &  +E  & Y & 85.0 & 84.2 & 84.6 & 74.9 	&	76.7  & 75.8 & 83.1 & 81.9 & 82.5\\
			\quad \textbf{+Tree-LSTM Syntax Encoder} & +E  & Y & 84.7  & 84.1 & 84.4  & 76.2  & 75.2 & 75.7 & 82.7 & 81.9 & 82.3 \\
			\midrule
			\textbf{Tree-based} & +E  & N & 85.4  & 83.6  & 84.5 & 76.1  & 75.1  & 75.6 & 83.3 & 81.9 & 82.6 \\
			\quad \textbf{+GCN Syntax Encoder} & +E  & Y & 84.5 & 85.9 & \bf 85.2 & 76.7 &	75.9  & \bf 76.3 & 82.9 &	83.3  & \bf 83.1 \\
			\quad \textbf{+SA-LSTM Syntax Encoder} &  +E  & Y & 85.0 & 85.0  & 85.0 & 77.2 & 75.0  & 76.1 &83.5  & 82.7  & \bf 83.1\\
			\quad \textbf{+Tree-LSTM Syntax Encoder} & +E  & Y & 85.7  & 84.1  & 84.9 & 75.5 & 75.9  & 75.7 & 83.0 & 82.4  & 82.7 \\
			\midrule
			\textbf{Graph-based} (\citeyear{li2019dependency}) & +E & N & 85.2 & 87.5 & 86.3 & 74.7  & 78.1 & 76.4 & 84.9 & 81.4 &  83.1 \\
			\quad\textbf{+Constituent Soft Pruning}  & +E  & Y & 87.1  & 85.7  & 86.4 & 77.0 &	76.2 & 76.6 & 83.4 	& 83.2 & 83.3  \\
			\quad \textbf{+GCN Syntax Encoder} & +E  & Y & 86.9 & 86.5  &  \bf 86.7 &77.5 & 76.3 & \bf 76.9 & 84.4  & 83.0 & \bf 83.7 \\
			\quad \textbf{+SA-LSTM Syntax Encoder} &  +E  & Y  & 87.3 &	85.7 & 86.5 & 76.0 & 77.2  & 76.6 & 83.8 &	83.2 & 83.5\\
			\quad \textbf{+Tree-LSTM Syntax Encoder} & +E  & Y & 85.8 & 86.6 & 86.2 & 76.9  & 76.1  & 76.5 & 83.6  & 82.8  & 83.2\\
			\bottomrule  
		\end{tabular}
	}
	
\end{table*}

We report the experimental results on the CoNLL-2005 and 2012 datasets without pre-identified predicates in Table \ref{tab:end-for-span}. Overall, our syntax-enhanced model using ELMo achieved the best $F_1$ scores on the CoNLL-2005 in-domain and CoNLL-2012 test sets. In comparison with the three baselines, our syntax utilization approaches consistently yielded better $F_1$ scores regardless of the factorization. Although the performance difference is small when using the constituent soft pruning on the graph-based model, the improvement seems natural because the constituent syntax for SRL has more to be explored than just boundary information.

\subsection{Dependency vs. Span}

\begin{table}
	\centering
	\caption{Dependency vs. Span-converted Dependency on the CoNLL-2005, CoNLL-2009 test sets with dependency evaluation.}\label{tab:convert}
	\setlength{\tabcolsep}{4pt}
	\begin{tabular}{clccc}
		\toprule
		& &  Dep F$_1$ & Span-converted F$_1$ & $\Delta$ F$_1$ \\
		\midrule
		\multirow{2}{25pt}{WSJ} & J \& N & 85.93 & 84.32 & 1.61 \\
		& Our system & 90.41 & 89.20 & 1.21 \\
		\hline
		\midrule
		\multirow{2}{25pt}{WSJ+\\Brown} & J \& N & 84.29 & 83.45 & 0.84 \\
		& Our system & 88.91 & 88.23 & 0.68 \\
		\bottomrule
	\end{tabular}
\end{table}

It is very hard to say which style of semantic formal representation, dependency or span, would be more convenient for machine learning as they adopt incomparable evaluation metrics. Recent research \cite{peng-etal-2018-learning} has proposed to learn semantic parsers from multiple datasets in Framenet style semantics, while our goal is to compare the quality of different models in span and dependency SRL for Propbank style semantics. Following \citet{johansson-nugues-2008-dependency}, we choose to directly compare their performance in terms of dependency-style metric through transformation. 
Using the head-finding algorithm in \cite{johansson-nugues-2008-dependency} which used gold-standard syntax, we may determine a set of head nodes for each span. This process will output an upper bound performance measure about the span conversion due to the use of gold syntax.

Based on our syntax-agnostic graph-based baseline, we do not train new models for the conversion and the resultant comparison. Instead, we use the span-style CoNLL 2005 test set and the dependency-style CoNLL 2009 test set (WSJ and Brown), considering these two test sets share the same text content. As the former only contains verbal predicate-argument structures, for the latter, we discard all nomial predicate-argument related results and predicate disambiguation results during performance statistics. Table \ref{tab:convert} shows the comparison.

On a more strict setting basis, the results from our same model for span and dependency SRL verify the same conclusion of \citet{johansson-nugues-2008-dependency}, namely, dependency form is more favorable in machine learning for SRL even compared to the conversion upper bound of the span form.

\subsection{Syntax Role under Different Pre-trained Langauge Models}

\begin{table}
	\centering
	\caption{SRL results with different pre-trained language models on CoNLL-2005, CoNLL-2009, and CoNLL-2012 test sets. The results listed in the table are evaluated while given pre-identified predicates. The baseline is our proposed graph-based model because of its good performance. The "SYN"  column indicates whether syntax information is employed, and the GCN syntax encoder is adopted for all models which are enhanced with syntax information. In this table, "$\triangle$" in brackets represent relative improvements using syntax information compared to the syntax-agnostic model, while the "$\uparrow$" indicates an absolute improvement using a pre-trained language model compared to the pure baseline model without any syntax information or pre-trained language model enhancement.}\label{tab:results_plm}
	\setlength{\tabcolsep}{3pt}
	\scalebox{0.75}{
		\begin{tabular}{lcccccc}
			\toprule  
			System & SYN & CoNLL09 WSJ & CoNLL09 Brown &  CoNLL05 WSJ & CoNLL05 Brown & CoNLL12\\
			\midrule
			\multirow{2}{*}{Baseline} & N & 87.8 & 79.2 & 84.5 &  74.8 &  83.5 \\
			& Y & 89.2 ($\triangle$ +1.4) & 80.1  ($\triangle$ +0.9)	& 85.6 ($\triangle$ +1.1)	  & 76.2 ($\triangle$ +1.4)	& 84.7 ($\triangle$ +1.2)  \\
			\midrule
			\multirow{2}{*}{ELMo} & N & 90.0 ($\ \uparrow\ $ +2.2) & 81.5 ($\ \uparrow\ $ +2.3) & 87.7 ($\ \uparrow\ $ +3.2) & 80.5 ($\ \uparrow\ $ +5.7) & 86.0  ($\ \uparrow\ $ +2.5)\\
			& Y & 91.1 ($\triangle$ +1.1)	& 82.1 ($\triangle$ +0.6)	& 88.6 ($\triangle$ +0.9)  &  81.0 	($\triangle$ +0.5)  & 86.7 	($\triangle$ +0.7) \\ \hdashline
			\multirow{2}{*}{BERT} & N & 91.4 ($\ \uparrow\ $ +3.6) & 82.8 ($\ \uparrow\ $ +3.6)	 & 89.0 ($\ \uparrow\ $ +4.5)	& 82.3 ($\ \uparrow\ $ +7.5)  & 87.4 ($\ \uparrow\ $ +3.9) \\
			& Y & 91.8 ($\triangle$ +0.4) & 83.2 ($\triangle$ +0.4) & 89.6 ($\triangle$ +0.6) & 82.8 ($\triangle$ +0.5) & 87.9 ($\triangle$ +0.5) \\ \hdashline
			\multirow{2}{*}{RoBERTa} & N & 91.4 ($\ \uparrow\ $ +3.6) & 83.1 	($\ \uparrow\ $ +3.9) & 89.3  ($\ \uparrow\ $ +4.8)	 & 82.7  ($\ \uparrow\ $ +7.9) & 87.9 	($\ \uparrow\ $ +4.4)  \\
			& Y & 91.7 ($\triangle$ +0.3)	& 83.2 ($\triangle$ +0.1)  & 89.7 ($\triangle$ +0.4) & 83.4  ($\triangle$ +0.7) & 88.0 	($\triangle$ +0.1)  \\ \hdashline
			\multirow{2}{*}{XLNet} & N & 91.5 ($\ \uparrow\ $ +3.7) & 84.1  ($\ \uparrow\ $ +4.9) & 89.8 	($\ \uparrow\ $ +5.3) &  85.2  ($\ \uparrow\ $ +10.4)  & 88.2 ($\ \uparrow\ $ +4.7)   \\
			& Y & 91.6 ($\triangle$ +0.1) & 84.2  ($\triangle$ +0.1) & 89.8 ($\triangle$ +0.0) & 85.4 ($\triangle$ +0.2) & 88.3 ($\triangle$ +0.1)   \\ \hdashline
			\multirow{2}{*}{ALBERT} & N & 91.6 ($\ \uparrow\ $ +3.8)	& 84.0 	($\ \uparrow\ $ +4.8) & 90.0 	($\ \uparrow\ $ +5.5)	& 84.9 ($\ \uparrow\ $ +10.1)	 & 88.5 ($\ \uparrow\ $ +5.0) \\
			& Y & 91.6 ($\triangle$ +0.0) & 84.3 ($\triangle$ +0.3)	 & 90.1 ($\triangle$ +0.1)  & 85.1	($\triangle$ +0.2)	& 88.7 ($\triangle$ +0.2)   \\
			\bottomrule
		\end{tabular}
	}
\end{table}

Language modeling, as an unsupervised natural language training technique, can produce a pre-trained language model by training generally on a large amount of text before further training on a more specific one out of a variety of natural language processing tasks. The downstream tasks then employ the obtained pre-trained models for further enhancement. After the introduction of pre-trained language models, they quickly and greatly dominated the performance of downstream tasks. Typical language models are ELMo \cite{peters-etal-2018-deep}, BERT \cite{devlin-etal-2019-bert}, RoBERTa \cite{liu2019roberta}, XLNet \cite{yang2019xlnet}, and ALBERT \cite{lan2019albert}, etc.. Therefore, in order to further explore the role of syntax in SRL on the strong basis of pre-trained language models, we evaluated the performance of these typical pre-trained language models on the best-performing combination of the graph-based baseline and GCN syntax encoder. The experimental results are shown in Table \ref{tab:results_plm}.

From the results in the table, all systems using the pre-trained language model have been greatly improved compared with the baseline models, especially in the out-of-domain test set. This indicates that since the pre-trained language model is trained on a very large scale corpus, the performance decrease caused by the domain inconsistency between the training and the test data is reduced, and the generalization ability for the SRL model is enhanced.

For the role of syntax, we found that the improvement from syntax enhancement in the baseline is greater than which systems gained from using the pre-trained language models. When the ability of pre-trained language models is strengthened, the performance growth brought by syntax gradually declines, and in some very strong language models, syntax even brings no further improvements for SRL. We suspect the reason is that the pre-trained language model can learn some syntactic information implicitly through unsupervised language modeling training \cite{hewitt-manning-2019-structural,clark-etal-2019-bert}, thus superimposing explicit syntactic information will not bring as much improvement as it will on the baseline model. We also found that the implicit features for pre-trained language models do not fully maximize syntactic information, as providing explicit syntactic information can still lead to improvements, though this depends on the quality of syntactic information and the ability to extract explicit syntactic information. When explicit syntax is accurate enough and the syntax feature encoder is strong enough,  the syntax information can further enhance the accuracy of SRL; however, if these conditions are not satisfied, then syntax is no longer the first option for promoting SRL performance in the neural network era. Besides, due to the uncontrollability of the implicit features of the pre-trained models, syntax as an auxiliary tool for SRL will exist for a long time, and neural SRL models can see even higher accuracy gains by leveraging syntactic information rather than ignoring it until neural networks' black box is fully revealed.

\subsection{Syntactic Contribution}

Syntactic information plays an informative role in semantic role labeling; however, few studies have been done to quantitatively evaluate the contribution of syntax to SRL. 
In dependency SRL, we observe that most of the neural SRL systems compared above used the syntactic parser of~\cite{bjorkelund-etal-2010-high} for syntactic inputs instead of the one from the CoNLL-2009 shared task, which adopted a much weaker syntactic parser. In particular,  \citet{marcheggiani-titov-2017-encoding} adopted an external syntactic parser with even higher parsing accuracy. Contrarily, our SRL model is based on the automatically predicted parse with moderate performance provided by the CoNLL-2009 shared task, but we still manage to outperform their models.
In span SRL, \citet{he-etal-2017-deep} injected syntax as a decoding constraint without having to retrain the model and compared the auto and gold syntax information for SRL. \citet{strubell-etal-2018-linguistically} presented a neural network model in which syntax is incorporated by training one attention head to attend to syntactic parents for each token and demonstrated that the SRL models benefit from injecting state-of-the-art predicted parses. Since different types of syntax and syntactic parsers are used in different works, the results are not directly comparable.
Thus, this motivates us to explore how much syntax contributes to dependency-based SRL in the deep learning framework and how to effectively evaluate the relative performance of syntax-based SRL. To this end, we conduct experiments for empirical analysis with different syntactic inputs.

\begin{algorithm}[t] 
	\caption{Faulty syntactic tree generator.} 
	\label{alg:sgd} 
	\begin{algorithmic}[1]
		\renewcommand{\algorithmicrequire}{\textbf{Input:}}
		\renewcommand{\algorithmicensure}{\textbf{Output:}}
		\REQUIRE A gold standard syntactic tree $GT$, the specific error probability $p$
		\ENSURE The new generative syntactic tree $NT$
		\STATE $N$ denotes the number of nodes in $GT$
		\FOR{each node $n \in GT$ }
		\STATE $r=$ random(0, 1), a random number
		\IF{$r<p$}
		\STATE $h=$ random(0, $N$), a random integer
		\STATE find the syntactic head $n_h$ of $n$ in $GT$
		\STATE modify $n_h=h$, and get a new tree $NT$
		\IF{$NT$ is a valid tree}
		\STATE \textbf{break}
		\ELSE
		\STATE \textbf{goto} step 5
		\ENDIF
		\ENDIF
		\ENDFOR
		\RETURN the new generative tree $NT$
	\end{algorithmic} 
\end{algorithm}

\paragraph{Syntactic Input}
In dependency SRL, four types of syntactic input are used to explore the role of syntax:
\begin{enumerate}
	\item The automatically predicted parse provided by CoNLL-2009 shared task.
	\item The parsing results of the CoNLL-2009 data by state-of-the-art syntactic parser, the Biaffine Parser \cite{dozat2017deep}.
	\item Corresponding results from another parser, the BIST Parser \cite{kiperwasser-goldberg-2016-simple}, which is also adopted by \citet{marcheggiani-titov-2017-encoding}.
	\item The gold syntax available from the official data set.
\end{enumerate}
Besides, to obtain flexible syntactic inputs for research, we design a faulty syntactic tree generator (referred to as STG hereafter) that is able to produce random errors in the output dependency tree like a real parser does. To simplify implementation, we construct a new syntactic tree based on the gold standard parse tree. Given an input error probability distribution estimated from a true parser output, our algorithm presented in Algorithm~\ref{alg:sgd} stochastically modifies the syntactic heads of nodes on the premise of a valid tree. In span SRL, since the dataset only provides a golden constituent syntax, we compared the auto syntax from the Choe\&Charniak Parser \cite{choe-charniak-2016-parsing}, Kitaev\&Klein Parser \cite{kitaev-klein-2018-constituency}, and the gold syntax from the dataset for SRL models.

\paragraph{Evaluation Measure}
For the SRL task, the primary evaluation measure is the semantic labeled F$_1$ score; however, this score is influenced by the quality of syntactic input to some extent, leading to an unfaithful reflection of the competence of a syntax-based SRL system. Namely, this is not the outcome of a true and fair quantitative comparison for these types of SRL models. To normalize the semantic score relative to syntactic parse, we take into account an additional evaluation measure to estimate the actual overall performance of SRL. Here, we use the ratio between labeled F$_1$ score for semantic dependencies (Sem-F$_1$) and the labeled attachment score (LAS) for syntactic dependencies proposed by~\citet{surdeanu-etal-2008-conll} as the evaluation metric in dependency SRL.\footnote{The idea of the ratio score in \cite{surdeanu-etal-2008-conll} actually was from an author of this paper, Hai Zhao; this was indicated in the acknowledgment part of \cite{surdeanu-etal-2008-conll}.} The benefits of this measure are twofold: it quantitatively evaluates syntactic contribution to SRL and impartially estimates the true performance of SRL, independent of the performance of the input syntactic parser. In addition, we further extended this evaluation metric for span SRL and used the ratio of Sem-F$_1$ and constituent syntax F$_1$ score (Syn-F$_1$) to measure the pure contribution of the proposed SRL model to clearly show the source of SRL performance improvement from the model's contribution rather than the improvement due to syntax accuracy.

\begin{table*}
	\centering
	\caption{Dependency SRL results on the  CoNLL 2009 English WSJ test set, in terms of labeled attachment score for syntactic dependencies (LAS), semantic precision (P), semantic recall (R), semantic labeled F$_1$ score (Sem-F$_1$), and the ratio Sem-F$_1$/LAS. A superscript "*" indicates LAS results from our personal communication with the authors.}\label{tab:ratio}
	\setlength{\tabcolsep}{4pt}
	\scalebox{0.75}{
		\begin{tabular}{l|c|cccccc}
			\toprule
			System & & PLM & LAS & P & R & Sem-F$_1$  & Sem-F$_1$/LAS \\
			\midrule
			\cite{zhao-etal-2009-multilingual} & &  & 86.0 & $-$ & $-$ & 85.4 & 99.30 \\
			\cite{zhao-etal-2009-multilingual-dependency} &  & & 89.2 & $-$ & $-$ & 86.2 & 96.64 \\
			\cite{bjorkelund-etal-2010-high} & &  & 89.8 & 87.1 & 84.5 & 85.8 & 95.55 \\
			\cite{lei-etal-2015-high} &  & & 90.4 & $-$ & $-$ & 86.6 & 95.80 \\
			\cite{fitzgerald-etal-2015-semantic} &  & & 90.4 & $-$ & $-$ & 86.7 & 95.90 \\
			\cite{roth-lapata-2016-neural} &  & & 89.8 & 88.1 & 85.3 & 86.7 & 96.5\\
			\cite{marcheggiani-titov-2017-encoding} & &  & 90.34$^*$ & 89.1 & 86.8 & 88.0 & 97.41 \\
			\midrule
			\multirow{3}{100pt}{Sequence-based + K-order hard pruning} &  CoNLL09 predicted & +E & 86.0 & 89.7 & 89.3 & 89.5 & 104.07 \\
			& STG Auto syntax &  +E & 90.0 & 90.5 & 89.3 & 89.9 & 99.89 \\
			& Gold syntax &  +E & 100.0 & 91.0 & 89.7 & 90.3 & 90.30 \\
			\midrule
			\multirow{4}{100pt}{Sequence-based + Syntax GCN encoder} &  CoNLL09 predicted &  +E & 86.0 & 90.5 & 88.5 & 89.5 & 104.07 \\
			& Biaffine Parser &  +E & 90.22 & 90.3 & 89.3 & 89.8 & 99.53 \\
			& BIST Parser &  +E & 90.05 & 90.3 & 89.1 & 89.7 & 99.61 \\
			& Gold syntax &  +E & 100.0 & 91.0 & 90.0 & 90.5 & 90.50 \\
			\bottomrule
		\end{tabular}
	}
\end{table*}

\begin{table*}
	\centering
	\caption{Span SRL results on the  CoNLL 2005 English WSJ test set, in terms of constituent syntax F$_1$ score (Syn-F$_1$), semantic precision (P), semantic recall (R), semantic labeled F$_1$ score (Sem-F$_1$), and the ratio Sem-F$_1$/Syn-F$_1$. }\label{tab:span_ratio}
	\setlength{\tabcolsep}{4pt}
	\scalebox{0.75}{
		\begin{tabular}{l|c|cccccc}
			\toprule
			System & & PLM & Syn-F$_1$ & P & R& Sem-F$_1$ & Sem-F$_1$/Syn-F$_1$ \\
			\midrule
			\multirow{2}{*}{\cite{he-etal-2017-deep}} & Choe\&Charniak Parser &  & 93.8 & $-$ & $-$ & 84.8 & 90.41 \\
			& Gold syntax & & 100.0 & $-$ & $-$ & 87.0 & 87.00 \\ \hdashline
			\multirow{2}{*}{\cite{wang2019best}} & Kitaev\&Klein Parser & +E & 95.4  & $-$ & $-$ & 88.2 & 92.45 \\
			& Gold syntax & +E  & 100.0 & $-$ & $-$ & 92.2 & 92.20 \\ \hdashline
			\cite{marcheggiani2019graph} &  Kitaev\&Klein Parser & & 95.4 & 85.8 & 85.1 & 85.4 & 89.52 \\
			\midrule
			\multirow{4}{100pt}{Sequence-based + Syntax GCN encoder}  & Choe\&Charniak Parser & & 93.8 & 86.4 & 84.8 & 85.6 & 91.26 \\
			& Choe\&Charniak Parser & +E & 93.8 & 87.2  & 86.8 &  87.0 & 92.75 \\
			& Kitaev\&Klein Parser & +E & 95.4 & 88.3 & 89.1 & 88.5 & 92.77 \\
			& Gold syntax & +E & 100.0 & 93.2 & 92.4 & 92.6 & 92.60 \\
			\bottomrule
		\end{tabular}
	}
\end{table*}

Table \ref{tab:ratio} reports the dependency SRL performance of existing models\footnote{Note that several SRL systems that do not provide syntactic information are not listed in the table.} in terms of Sem-F$_1$/LAS ratio on the CoNLL-2009 English test set. Interestingly, even though our system has significantly lower scores than others by 3.8\% LAS in syntactic components, we obtain the highest results on both Sem-F$_1$ and Sem-F$_1$/LAS ratio. These results show that our SRL component is relatively much stronger. Moreover, the ratio comparison in Table \ref{tab:ratio} also shows that since the CoNLL-2009 shared task, most SRL works actually benefit from the enhanced syntactic component rather than the improved SRL component itself. No post-CoNLL SRL systems, neither traditional nor neural types, exceeded the top systems of the CoNLL-2009 shared task, \cite{zhao-etal-2009-multilingual} (SRL-only track using the provided predicated syntax) and \cite{zhao-etal-2009-multilingual-dependency} (Joint track using self-developed parser). We believe that this work, for the first time, reports both a higher Sem-F$_1$ and a higher Sem-F$_1$/LAS ratio since the CoNLL-2009 shared task. We also presented the comprehensive results of our sequence-based SRL model with a syntax encoder instead of pruning on the aforementioned syntactic inputs of different quality and compare these with previous SRL models. A number of observations can be made from these results. First, the model with the GCN syntax encoder gives quite stable SRL performance no matter the syntactic input quality, which varies in a broad range, and it obtains overall higher scores compared to the previous states-of-the-art. Second, it is interesting to note that the Sem-F$_1$/LAS score of our model becomes relatively smaller as the syntactic input becomes better. Not surprisingly though, these results show that our SRL component is relatively even stronger. Third, when we adopt a syntactic parser with higher parsing accuracy, our SRL system achieves a better performance. Notably, our model yields a Sem-F$_1$ of 90.53\% taking gold syntax as input. This suggests that high-quality syntactic parse may indeed enhance SRL, which is consistent with the conclusion in \cite{he-etal-2017-deep}. 

We further evaluated the syntactic contribution in span SRL, as shown in Table \ref{tab:span_ratio}. For \cite{wang2019best} and our results, when syntax and pre-trained language model were kept the same, our model obtained better sem-F$_1$, which is also reflected in the sem-F$_1$/syn-F$_1$ ratio, indicating that our model has stronger syntactic utilization ability. In addition, by comparing the results of whether a pre-trained language model is used, it can be found that those using the pre-trained language model have a higher ratio of Sem-F$_1$/Syn-F$_1$, which shows that the features offered by the pre-trained language model potentially increase the syntactic information, resulting in a higher ratio of syntax contribution.

\begin{figure}
	\centering
	\includegraphics[scale=0.52]{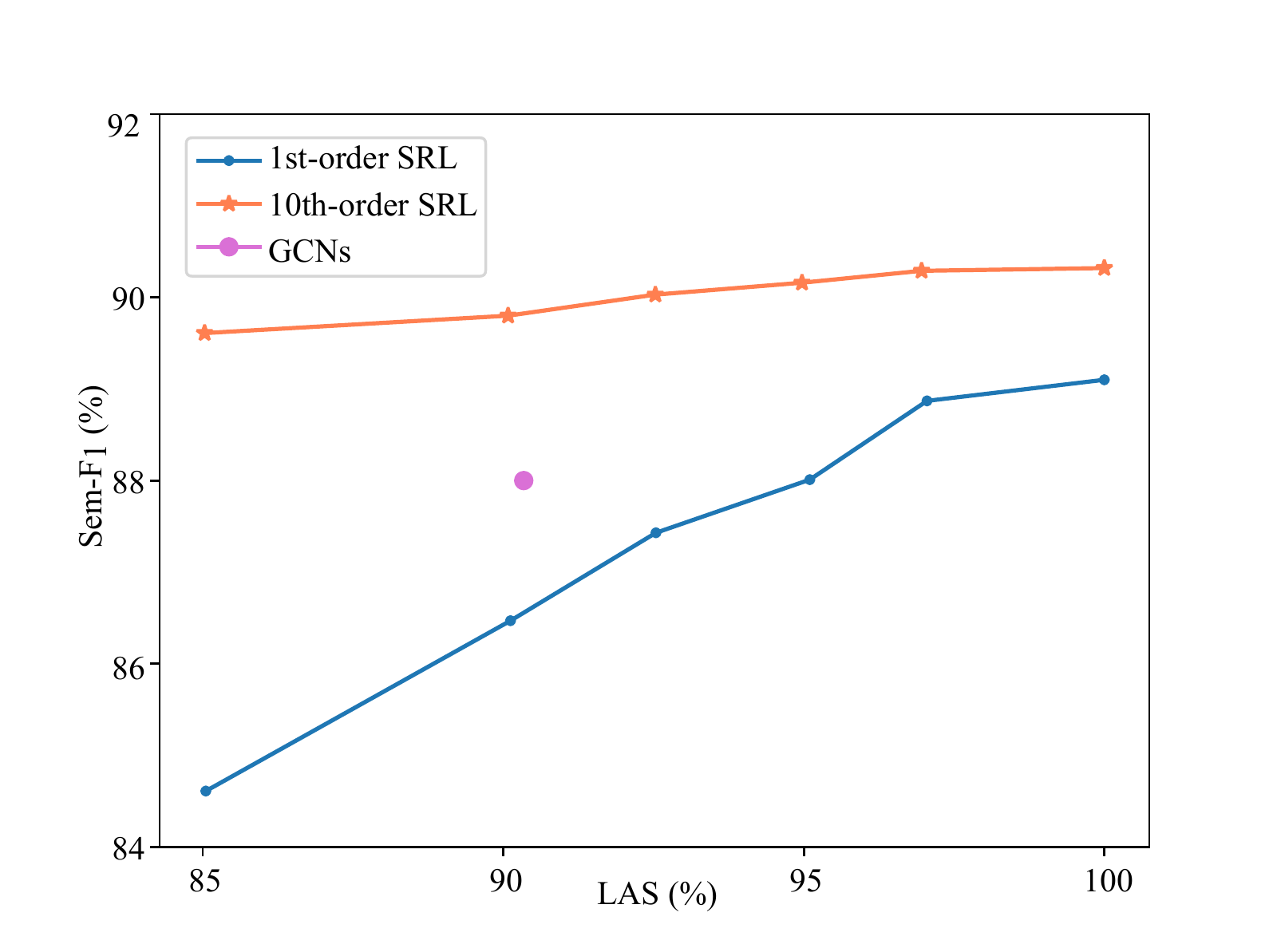}
	\caption{The Sem-F$_1$ scores of our models with different quality of syntactic inputs vs. original GCNs~\cite{marcheggiani-titov-2017-encoding} on CoNLL-2009 WSJ test set.} \label{fig:las}
\end{figure}

Besides, to show how SRL performance varies with syntax accuracy, we also test our sequence-based dependency SRL model with k-order hard pruning in first and tenth orders using different erroneous syntactic inputs generated from STG and evaluate their performance using the Sem-F$_1$/LAS ratio. Figure \ref{fig:las} shows Sem-F$_1$ scores at different qualities of syntactic parse inputs on the English test set, whose LAS varies from 85\% to 100\%. Compared to previous states-of-the-art~\cite{marcheggiani-titov-2017-encoding}. Our tenth-order pruning model gives quite stable SRL performance no matter the syntactic input quality, which varies in a broad range, while our first-order pruning model yields overall lower results (1-5\% F$_1$ drop), owing to missing too many true arguments. These results show that high-quality syntactic parses may indeed enhance dependency SRL. Furthermore, they indicate that our model with syntactic input as accurate as \citet{marcheggiani-titov-2017-encoding}, namely, 90\% LAS, will give a Sem-F$_1$ exceeding 90\%.

\section{Related Work}

Semantic role labeling was pioneered by~\citet{gildea-jurafsky-2000-automatic}. In dependency SRL, most traditional SRL models rely heavily on feature templates~\cite{pradhan-etal-2005-semantic,zhao-etal-2009-semantic,bjorkelund-etal-2009-multilingual}. Among them,~\citet{pradhan-etal-2005-semantic} combined features derived from different syntactic parses based on SVM classifier, while~\citet{zhao-etal-2009-semantic} presented an integrative approach for dependency SRL via a greedy feature selection algorithm. Later, ~\citet{collobert2011natural} proposed a convolutional neural network model that induced word embeddings rather than relied on handcrafted features, which was a breakthrough for the SRL task.

\citet{foland-martin-2015-dependency} presented a dependency semantic role labeler using convolutional and time-domain neural networks, while~\citet{fitzgerald-etal-2015-semantic} exploited neural networks to jointly embed arguments and semantic roles, akin to the work~\cite{lei-etal-2015-high} that induced a compact feature representation by applying a tensor-based approach. Recently, researchers have considered multiple ways to effectively integrate syntax into SRL learning. \citet{roth-lapata-2016-neural} introduced dependency path embedding to model syntactic information and exhibited notable success. \citet{marcheggiani-titov-2017-encoding} leveraged the graph convolutional network to incorporate syntax into neural models. Differently,~\citet{marcheggiani-etal-2017-simple} proposed a syntax-agnostic for dependency SRL that used effective word representation, which for the first time achieved performance comparable to state-of-the-art syntax-aware SRL models.

However, most neural SRL works seldom pay much attention to the impact of input syntactic parse over the resulting SRL performance. This work is thus more than proposing a high performance SRL model through reviewing the highlights of previous models; it also presents an effective syntax tree-based method for argument pruning. Our work is also closely related to \cite{punyakanok-etal-2008-importance}. Under traditional methods, \citet{punyakanok-etal-2008-importance} investigated the significance of syntax to SRL systems and showed syntactic information was most crucial in the pruning stage. There are two important differences between \cite{punyakanok-etal-2008-importance} and ours. First, in our paper, we summarize the current dependency and span SRL and consider them under multiple baseline models and syntax integration approaches to reduce deviations resulting from model structure and syntax integration approach. Second, the development of pre-trained language models has dramatically changed the basis of SRL models, which motivated us to revisit the role of syntax based on new situations.

In the other span SRL research lines, \citet{moschitti-etal-2008-tree} applied tree kernels as encoders to extract constituency tree features for SRL, while \citet{naradowsky-etal-2012-improving} used graphical models to model the tree structures. \citet{socher-etal-2013-recursive, tai-etal-2015-improved} proposed recursive neural networks for incorporating syntax information in SRL that recursively encoded constituency trees to constituent representations. \citet{he-etal-2017-deep} presented an extensive error analysis with deep learning models for span SRL and included a discussion of how constituent syntactic parsers could be used to improve SRL performance. With the recent advent of self-attention, syntax can be used not only for pruning or encoding to provide auxiliary features, but also for guiding the structural learning of models. \citet{strubell-etal-2018-linguistically} modified the dependency tree structure to train one attention head to attend to syntactic parents for each token. In addition, compared with the application of dependency syntax, the difficult step is mapping the constituent features back to the word level. \citet{wang2019best} extended a syntax linearization approaches of \citet{gomez-rodriguez-vilares-2018-constituent} and incorporated this information as a word-level feature in a SRL model; \citet{marcheggiani2019graph} introduced a novel neural architecture, SpanGCN, for encoding constituency syntax at the word level.

\section{Conclusion}

This paper explores the role of syntax for the semantic role labeling task. We presented a systematic survey based on our recent works on SRL and a recently popular pre-trained language modeling. Through experiments on both the dependency and span formalisms, and the sequence-based, tree-based and graph-based modeling approaches, we conclude that although the effects of syntax on SRL seem like a never-ending topic of research, with the help of current unsupervised pre-trained language models, the syntax improvement provided to SRL model performance seems to be gradually reaching its upper limit. Beyond presenting approaches that lead to improved SRL performances, we performed a detailed and fair experimental comparison between span and dependency SRL formalisms to show which is more fit for machine learning. In addition, we have studied a variety of methods of syntax integration and have shown that there is unacclimation for the hard pruning in the deep learning model which is very popular in the pre-NN era.



\section{Acknowledgments}

We thank Rui Wang (wangrui.nlp@gmail.com), from the National Institute of Information and Communications Technology (NICT), for his helpful feedback and discussions. We also thank Kevin Parnow (parnow@sjtu.edu.cn), from the Department of Computer Science and Engineering Shanghai Jiao Tong University for his kindly helps in proofreading when we were working on this paper. 

\starttwocolumn
\bibliography{compling_style}

\end{document}